\documentclass[10pt,twocolumn,letterpaper]{article}

\usepackage[pagenumbers]{cvpr}

\usepackage{amsfonts}
\usepackage{algorithm}
\usepackage{algorithmic}
\usepackage{array}
\usepackage{multirow}
\usepackage{siunitx}
\usepackage{textcomp}
\usepackage{placeins}
\usepackage{microtype}
\newcommand{\NA}{--}

\makeatletter
\patchcmd{\@maketitle}{\vskip .375in}{\vskip .125in}{}{\PackageError{arxivlayout}{Title top-spacing patch failed}{}}
\patchcmd{\@maketitle}{\vspace*{12pt}}{\vspace*{6pt}}{}{\PackageError{arxivlayout}{Title bottom-spacing patch failed}{}}
\patchcmd{\@maketitle}{\vspace*{24pt}}{\vspace*{12pt}}{}{\PackageError{arxivlayout}{Title author-spacing patch failed}{}}
\makeatother

\usepackage{needspace}

\definecolor{cvprblue}{rgb}{0.21,0.49,0.74}
\usepackage[pagebackref,breaklinks,colorlinks,allcolors=cvprblue]{hyperref}

\def\confName{CVPR}
\def\confYear{2027}

\title{Rescuing Discarded Boundary Samples in Semi-Supervised Learning}
\author{%
Yuzhao Jiang\quad Meng Zhang\thanks{Corresponding author.}\quad Jie Ma\quad Xiaohong Guan\\
Xi'an Jiaotong University, Xi'an, China
}
\hypersetup{
  pdftitle={Rescuing Discarded Boundary Samples in Semi-Supervised Learning},
  pdfauthor={Yuzhao Jiang, Meng Zhang, Jie Ma, Xiaohong Guan},
  pdfsubject={Semi-supervised learning},
  pdfkeywords={semi-supervised learning, pseudo-labeling, boundary refinement}
}

\begin{document}
\maketitle

% ===========================================================
\begin{abstract}
    The field of semi-supervised learning (SSL) has seen substantial progress,
    driven by ongoing refinements within the pseudo-label SSL paradigm. The
    main difficulty lies in reducing the adverse impact of
    confirmation bias. The predominant approach is to apply a confidence
    threshold during training to discard low-confidence samples. Nevertheless,
    while this mitigates the bias, the discarded samples are mostly situated near
    class boundaries, and consequently, this strategy also loses valuable
    boundary information. To address this problem, we propose \textbf{Boundary
    Refinement and Correction (BRC)}, an add-on for confidence-thresholded
    SSL that rescues useful discarded samples without changing how the base SSL
    method selects pseudo-labels. BRC keeps the original confidence-based
    selection rule and operates only on the
    rejected set.
    It first identifies discarded samples that may still be useful for training,
    then replaces their original pseudo-labels with corrected soft pseudo-labels
    and returns them to the unsupervised loss. Extensive experiments on
    CIFAR-100, STL-10, and TissueMNIST demonstrate the flexibility and broad
    applicability of BRC across diverse datasets and base SSL methods, reducing
    error rates by up to 4.97 percentage points. On boundary-near samples, BRC
    further improves Top-1 accuracy by 12.80 percentage points, directly
    supporting its ability to refine the model's decision boundary. An exploratory
    extension to semantic segmentation also yields consistent gains of
    $1.03$--$2.39$ mIoU, indicating that BRC generalizes beyond image
    classification.
\end{abstract}

% ===========================================================
\section{Introduction}
Semi-supervised learning (SSL) is an important part of deep learning and has
achieved remarkable success across a wide range of
tasks~\cite{cheng2025cgmatch,mai2026hess}. In pseudo-label-based SSL,
incorrect pseudo-labels can reinforce the model's existing mistakes. As a
result, these errors may persist or accumulate during subsequent training.
This phenomenon is known as confirmation bias~\cite{arazo2020pseudo}.
Confirmation bias is one of the central challenges in SSL.

\begin{figure}[t]
\centering
\includegraphics[width=0.9\columnwidth]{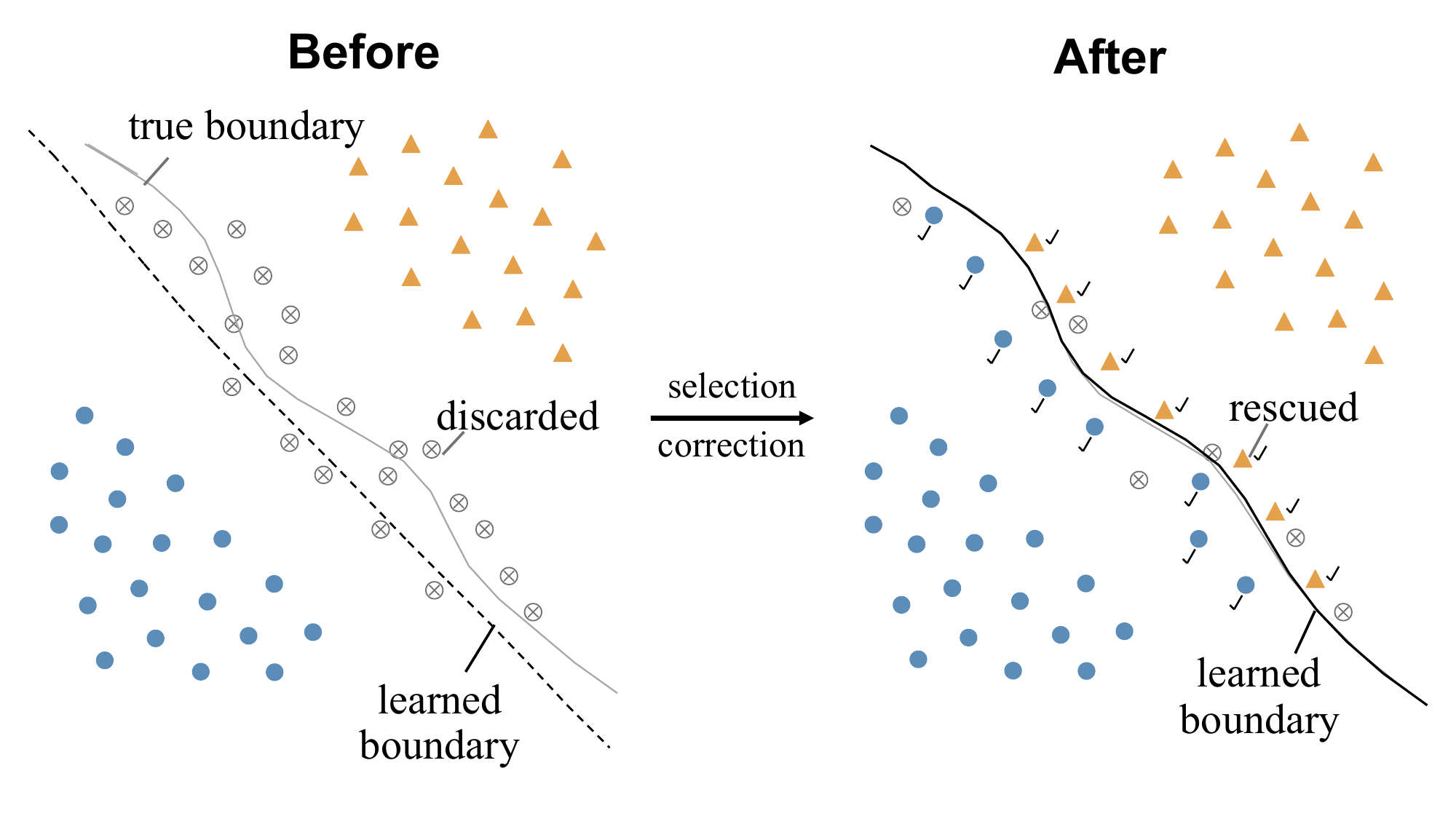}
\caption{Motivation of BRC.
Confidence thresholding removes samples near the class boundary and weakens
margin supervision. BRC selects and corrects useful discarded samples to
restore boundary information.}
\label{fig:brc_overview}
\end{figure}

To mitigate confirmation bias, a series of confidence-thresholded SSL methods
have been proposed, including
UDA~\cite{xie2020self}, FixMatch~\cite{sohn2020fixmatch},
FlexMatch~\cite{zhang2021flexmatch}, and
FreeMatch~\cite{wang2023freematch}. Their core idea is to use prediction
confidence as a measure of pseudo-label reliability. Only pseudo-labels that
meet the confidence threshold are retained for unsupervised training, while
the remaining samples are discarded in the current iteration. However, as
shown on the left of Figure~\ref{fig:brc_overview}, samples discarded by the
threshold often lie near class boundaries~\cite{zeng2023pefat}. While
confidence filtering removes unreliable pseudo-labels, it also discards many
samples that are important for learning the classifier margin~\cite{zhao2026iasft}. This leaves the
region near the boundary with insufficient supervision, so the learned
decision boundary may deviate from the true one. In summary, confidence
thresholding helps suppress pseudo-label noise, but may also lose valuable
information for learning the decision boundary.

To address this issue, we propose \textbf{Boundary Refinement and Correction
(BRC)}, an add-on module for confidence-thresholded SSL. BRC keeps
the base method's thresholding rule unchanged. Rather than treating discarded
samples as useless data, the method regards them as a boundary candidate pool.
For each discarded sample, BRC first estimates its potential training value for
the backbone. Based on these estimates, the method selects useful candidates,
corrects their unreliable pseudo-labels into more accurate soft targets, and
returns the selected samples to training. This preserves the noise-control
benefit of thresholding while rescuing informative boundary samples to refine
the decision boundary.

Extensive experiments on CIFAR-100~\cite{krizhevsky2009cifar},
STL-10~\cite{coates2011stl10}, and TissueMNIST~\cite{medmnist2021} demonstrate the
flexibility and broad applicability of BRC across diverse datasets and base SSL
methods, reducing error rates by up to 4.97 percentage points. On boundary-near
samples, BRC further improves Top-1 accuracy by 12.80 percentage points,
directly supporting its ability to refine the model's decision boundary. An
exploratory extension to semantic segmentation also yields consistent gains of
$1.03$--$2.39$ mIoU, indicating that BRC generalizes beyond image
classification.

The main contributions of this work are as follows.
\begin{itemize}
  \item We find that confidence-thresholded SSL discards many boundary-near
  samples together with unreliable pseudo-labels, thereby losing supervision
  important for learning the decision margin. We formulate the solution as a
  post-selection problem for selective rescue of discarded samples.

  \item We propose BRC, an add-on module that preserves the base SSL method's
  selection rule while estimating rescue utility, correcting unreliable
  pseudo-labels, and returning selected discarded samples to training.

  \item We integrate BRC with multiple confidence-thresholded SSL methods and
  evaluate it through main comparisons, representative component ablations, and
  related analyses. The results demonstrate the effectiveness of BRC and show that its error-rate reductions come from correcting and selectively reusing discarded boundary samples.

\end{itemize}

% ===========================================================
\section{Related work}

\subsection{Pseudo-label SSL}
Pseudo-labeling~\cite{lee2013pseudo} trains on model predictions as temporary
labels for unlabeled examples. Prior work mainly improves target quality and
reliability through prediction averaging, sharpening~\cite{berthelot2019mixmatch}, distribution alignment~\cite{berthelot2020remixmatch},
confidence filtering~\cite{sohn2020fixmatch}, and MixUp-style
regularization~\cite{berthelot2019mixmatch}. Open-set
SSL further shows that pseudo-labels become harder to trust when unlabeled data
are less constrained~\cite{han2026paf}. Other approaches improve pseudo-label
reliability through semantic and instance similarity~\cite{zheng2022simmatch,zhou2023hypermatch}
or training-dynamics-based pseudo-margins~\cite{sosea2023marginmatch}.
These works motivate BRC's focus on pseudo-label reliability. BRC differs by targeting samples that the base SSL
rule has already discarded.

In dense prediction, U2PL uses unreliable pixels as negative samples~\cite{wang2022u2pl},
UniMatch strengthens weak-to-strong consistency~\cite{yang2023unimatch}, and CorrMatch propagates
supervision through pixel and region correlations~\cite{sun2024corrmatch}. These studies show the importance of using
unlabeled pixels beyond conventional high-confidence pseudo-labels and motivate
our exploratory transfer of BRC to semantic segmentation.

\subsection{Confirmation bias and threshold-based methods}
Confirmation bias arises when incorrect pseudo-labels reinforce the model's
prediction errors~\cite{arazo2020pseudo}. Confidence-based filtering helps
mitigate this problem by excluding unreliable targets~\cite{wu2023chmatch,cheng2025cgmatch,vishwakarma2025rethinking,liu2026covar}.
FixMatch~\cite{sohn2020fixmatch}, FlexMatch~\cite{zhang2021flexmatch}, and
FreeMatch~\cite{wang2023freematch} use fixed, class-adaptive, and self-adaptive
thresholds, respectively. Subsequent methods improve confidence use through
soft weighting~\cite{chen2023softmatch}, low-confidence MixUp~\cite{han2025regmixmatch},
or optimal-transport regularization~\cite{tan2024otmatch}.
These threshold-strategy methods serve as natural base methods to which BRC can be attached.

\subsection{Meta-learning}
Meta-learning learns across tasks to support adaptation from limited
data~\cite{finn2017maml,nichol2018firstorder}, with recent work addressing
computational scalability~\cite{zhang2023ibaml}, generalization
guarantees~\cite{wang2024metastability}, and task-specific
regularization~\cite{iwata2025regularization}.
Meta Pseudo Labels~\cite{pham2021meta} applies meta-learning to
semi-supervised learning.
However, our experiments show that, for an SSL learner with adaptive
confidence selection, meta-learned correction alone fails to improve
performance and can even degrade it. Combining selection based on utility scores
with correction yields the strongest gains, as demonstrated in
Supplementary Secs.~5.1--5.2.

% ===========================================================
\section{Preliminary}
\label{sec:preliminary}

% Place the full-width framework early enough to appear near its first detailed
% discussion in the Method overview, rather than one page later.
\begin{figure*}[t]
\centering
\includegraphics[width=\textwidth]{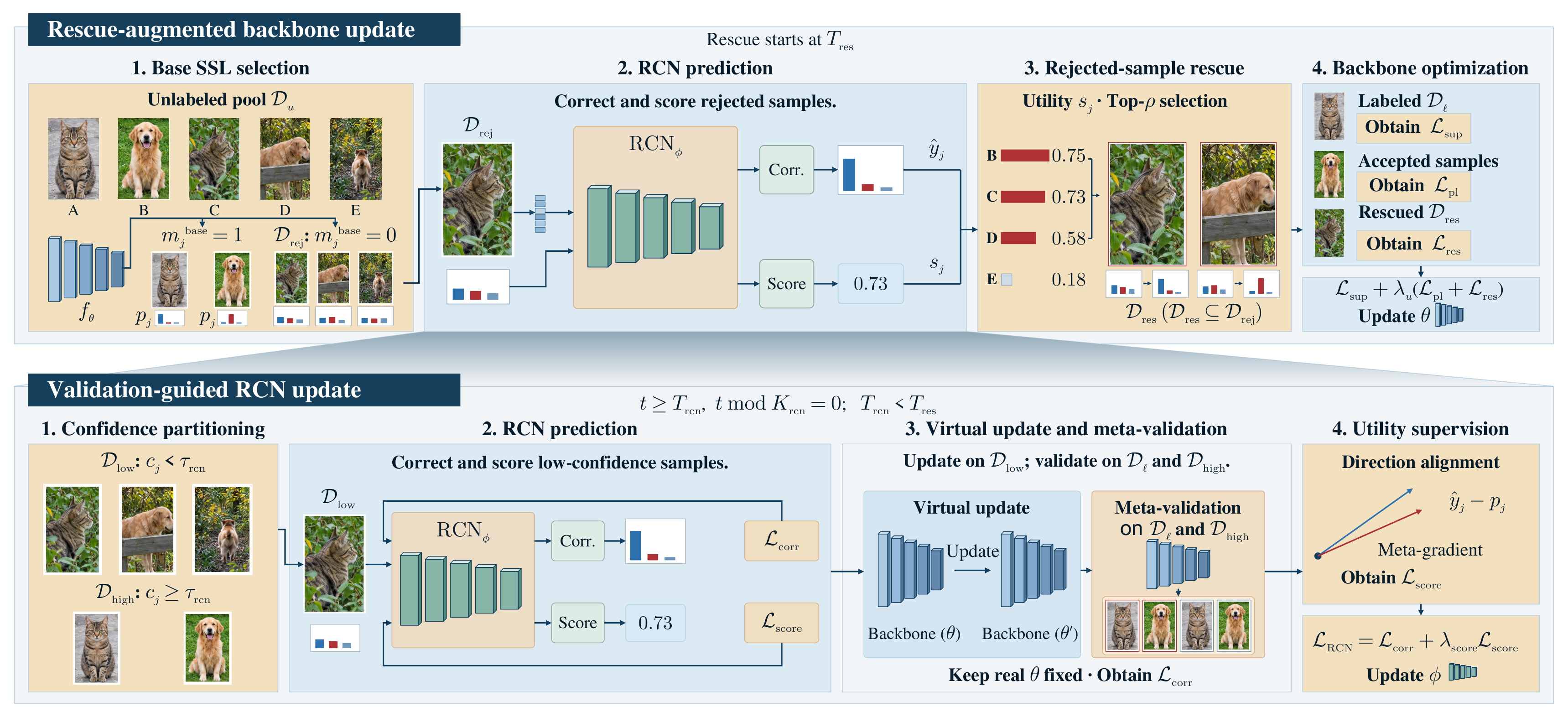}
\caption{Overall BRC framework. Top: the backbone update selects
useful samples from $\mathcal{D}_{\mathrm{rej}}$, assigns corrected targets
$\widehat{y}_j$, and reuses them through $\mathcal{L}_{\mathrm{res}}$.
Bottom: the RCN update uses virtual updates and gradient-direction
comparison to learn corrections and utility scores.}
\label{fig:brc_framework}
\end{figure*}

\noindent\textbf{Confidence-Thresholded Pseudo-Label SSL.}
We consider labeled data
$\mathcal{D}_{\ell}=\{(x_i,y_i)\}_{i=1}^{B_{\ell}}$ and unlabeled data
$\mathcal{D}_u=\{u_j\}_{j=1}^{B_u}$. At each iteration, the base SSL model
predicts a class-probability distribution $p_j$ for each unlabeled sample
$u_j$, yielding a pseudo-label $\widetilde{y}_j$ and confidence $c_j$. The base
SSL method uses $c_j$ together with
its own thresholding rule to produce the binary selection mask $m_j$. This
single mask covers fixed-threshold,
class-adaptive, and self-adaptive selection rules.

The backbone is updated using the supervised loss
$\mathcal{L}_{\mathrm{sup}}$ and the pseudo-label loss
$\mathcal{L}_{\mathrm{pl}}$.
The labeled data provide the supervised loss
\begin{equation}
    \mathcal{L}_{\mathrm{sup}}
    = \frac{1}{B_{\ell}}\sum_{i=1}^{B_{\ell}}
      \mathrm{CE}\!\left(f_\theta(x_i),y_i\right),
    \label{eq:sup_loss}
\end{equation}
while the selected unlabeled samples provide the pseudo-label loss
\begin{equation}
    \mathcal{L}_{\mathrm{pl}}
    = \frac{1}{B_u}\sum_{j=1}^{B_u}
      m_j\,\mathrm{CE}\!\left(f_\theta(u_j),\widetilde{y}_j\right),
    \label{eq:pl_loss}
\end{equation}
where $f_\theta$ is the backbone classifier with parameters $\theta$, and
$\mathrm{CE}$ denotes cross-entropy. The coefficient $\lambda_u$ weights the
unlabeled loss. Samples with $m_j=0$ are discarded by the base SSL selection rule in the
current iteration and form the rejected set $\mathcal{D}_{\mathrm{rej}}$.

\noindent\textbf{BRC Extension.}
BRC preserves the base method's selection mask and pseudo-label loss on
accepted samples. It processes only $\mathcal{D}_{\mathrm{rej}}$. From this
set, BRC selects a useful subset
$\mathcal{D}_{\mathrm{res}}\subseteq\mathcal{D}_{\mathrm{rej}}$. BRC corrects
the pseudo-labels of samples in $\mathcal{D}_{\mathrm{res}}$ and reintroduces
them into backbone updates through an additional rescue loss.
% ===========================================================
\section{Method}
\label{sec:method}

\subsection{Overview}
\label{subsec:method_overview}

BRC augments the base SSL process with a Refinement and Correction Network
(RCN), denoted by $\mathrm{RCN}_\phi$, where $\phi$ denotes its parameters.
For each rejected sample, the RCN estimates its training utility and corrects
its original pseudo-label. Figure~\ref{fig:brc_framework} summarizes the
\emph{rescue-augmented backbone update} and \emph{validation-guided RCN update}.
During backbone updates, the RCN parameters remain fixed.
At each iteration, the base SSL method first partitions the unlabeled samples
into accepted and rejected sets using its original confidence-based selection
rule. The RCN then selects useful rejected samples, corrects their pseudo-labels,
and reintroduces them into backbone updates. During RCN updates, the real backbone
parameters $\theta$ remain fixed. A virtual backbone update measures the effect
of the current RCN outputs on reliable data. This feedback guides the RCN update.

\subsection{Rescue-augmented backbone update}
\label{subsec:backbone_update}

The backbone update augments the base SSL objective with useful supervision
from rejected samples while preserving the update path for accepted samples. This rescue path is activated at
$T_{\mathrm{res}}$; for $t<T_{\mathrm{res}}$, the backbone follows the standard
SSL update.

The RCN parameters remain fixed during the backbone update. Each
$u_j\in\mathcal{D}_{\mathrm{rej}}$ is processed by the RCN, which outputs a
corrected soft pseudo-label $\widehat{y}_j$ and a utility score $s_j$ from its
score head. BRC
ranks the rejected samples by $s_j$ and selects the top $\rho$ proportion to
form $\mathcal{D}_{\mathrm{res}}$. Only the selected samples are assigned
corrected pseudo-labels and reintroduced into backbone updates. Here,
$0<\rho\leq1$ is the rescue ratio.

After obtaining $\mathcal{D}_{\mathrm{res}}$, the rescue loss $\mathcal{L}_{\mathrm{res}}$
is
\begin{equation}
    \mathcal{L}_{\mathrm{res}}
    = \frac{1}{|\mathcal{D}_{\mathrm{res}}|}
      \sum_{u_j\in\mathcal{D}_{\mathrm{res}}}
      \mathrm{CE}\!\left(
      f_\theta(u_j),\widehat{y}_j
      \right).
    \label{eq:rescue_loss}
\end{equation}
Here, $|\cdot|$ denotes set cardinality.
When $\mathcal{D}_{\mathrm{res}}=\varnothing$, we set
$\mathcal{L}_{\mathrm{res}}=0$.
Once sample rescue is activated,
$\mathcal{L}_{\mathrm{sup}}$, $\mathcal{L}_{\mathrm{pl}}$, and
$\mathcal{L}_{\mathrm{res}}$ define the backbone objective:
\begin{equation}
    \mathcal{L}_{\mathrm{BRC}}
    = \mathcal{L}_{\mathrm{sup}}
    + \lambda_u\left(\mathcal{L}_{\mathrm{pl}}
    + \mathcal{L}_{\mathrm{res}}\right).
    \label{eq:main_loss}
\end{equation}

\subsection{Validation-guided RCN update}
\label{subsec:rcn_training}

The RCN is updated to learn reliable pseudo-label corrections and estimate their
training utility. RCN updates start at iteration $T_{\mathrm{rcn}}$ and are performed every
$K_{\mathrm{rcn}}$ iterations thereafter. The condition
$T_{\mathrm{rcn}}<T_{\mathrm{res}}$ provides an initial RCN training stage
before sample rescue begins.

The RCN is a dual-head network with a correction head and a score head. It
takes the backbone feature and predictive distribution $p_j$ as input.
The correction head outputs the corrected soft pseudo-label $\widehat{y}_j$,
while the score head outputs the utility score $s_j\in(0,1)$.

Because base methods may use different selection rules, the RCN update uses an
independent confidence threshold
$\tau_{\mathrm{rcn}}$. Samples with $c_j<\tau_{\mathrm{rcn}}$ form the low-confidence set
$\mathcal{D}_{\mathrm{low}}$, while those with
$c_j\geq\tau_{\mathrm{rcn}}$ form the high-confidence set
$\mathcal{D}_{\mathrm{high}}$. The low-confidence set provides candidate
corrections for a virtual backbone update. The labeled data and
$\mathcal{D}_{\mathrm{high}}$ provide reliable meta-validation data. We apply
one virtual backbone update using the corrected labels from
$\mathcal{D}_{\mathrm{low}}$. The virtual backbone is then evaluated on
$\mathcal{D}_{\ell}$ and $\mathcal{D}_{\mathrm{high}}$. This validation
feedback trains the correction head and provides a binary target for the score
head. The target indicates whether each proposed correction is beneficial.

The RCN first produces $\widehat{y}_j$ and $s_j$ for each sample in
$\mathcal{D}_{\mathrm{low}}$.
Using the corrected pseudo-labels, we take one differentiable virtual step from
$\theta$ and obtain the virtual backbone parameters $\theta'$:
\begin{equation}
    \theta' = \theta-\alpha\nabla_\theta
    \left[
    \frac{1}{|\mathcal{D}_{\mathrm{low}}|}
    \sum_{u_j\in\mathcal{D}_{\mathrm{low}}}
    \mathrm{CE}\!\left(f_\theta(u_j),\widehat{y}_j\right)
    \right],
    \label{eq:virtual_update}
\end{equation}
where $\alpha$ is the virtual-update step size. The virtual parameters
$\theta'$ are used only inside the RCN update. They never replace the real
backbone parameters $\theta$.

The virtual backbone is evaluated on the labeled and high-confidence data.
The resulting meta-validation loss is
\begin{equation}
    \begin{aligned}
    \mathcal{L}_{\mathrm{corr}}
    ={}& \frac{1}{B_{\ell}+|\mathcal{D}_{\mathrm{high}}|}
      \Bigg[
      \sum_{i=1}^{B_{\ell}}\mathrm{CE}\!\left(f_{\theta'}(x_i),y_i\right) \\
      &+\sum_{u_j\in\mathcal{D}_{\mathrm{high}}}
      \mathrm{CE}\!\left(f_{\theta'}(u_j),\widetilde{y}_j\right)
      \Bigg].
    \end{aligned}
    \label{eq:correction_loss}
\end{equation}
The correction head is trained by minimizing
$\mathcal{L}_{\mathrm{corr}}$ through the virtual update. This objective
encourages the corrected targets to produce a backbone update that improves
performance on reliable data.

The score head requires an explicit target for each proposed correction. The
vector from $p_j$ to $\widehat{y}_j$ represents the correction proposed by the
RCN. The meta-gradient indicates the target-space direction that reduces the
validation error. When these two directions align, the correction is considered
beneficial. We therefore define the binary supervision target
$t_j$ as
\begin{equation}
    t_j = \mathbb{I}\!\left[
    -\left\langle
    \nabla_{\widehat{y}_j}\mathcal{L}_{\mathrm{corr}},
    \widehat{y}_j-p_j
    \right\rangle>0
    \right]\in\{0,1\},
    \label{eq:target}
\end{equation}
where $\mathbb{I}[\cdot]$ denotes the indicator function and
$\langle\cdot,\cdot\rangle$ denotes the inner product.
We train the score head with
\begin{equation}
    \mathcal{L}_{\mathrm{score}}
    = \frac{1}{|\mathcal{D}_{\mathrm{low}}|}
      \sum_{u_j\in\mathcal{D}_{\mathrm{low}}}
      \mathrm{BCE}(s_j,t_j),
    \label{eq:score_loss}
\end{equation}
where $\mathrm{BCE}$ denotes binary cross-entropy. The complete RCN objective
contains two terms. The meta-validation signal trains the correction head, and
the binary supervision trains the score head:
\begin{equation}
    \mathcal{L}_{\mathrm{RCN}}
    = \mathcal{L}_{\mathrm{corr}}
    + \lambda_{\mathrm{score}}\mathcal{L}_{\mathrm{score}},
    \label{eq:rcn_update}
\end{equation}
where $\lambda_{\mathrm{score}}$ is the score-loss weight. Minimizing
$\mathcal{L}_{\mathrm{RCN}}$ updates $\phi$ only. If
$\mathcal{D}_{\mathrm{low}}=\varnothing$, the RCN update is skipped.

Algorithm~\ref{alg:brc_training} summarizes how the backbone and RCN updates
are scheduled and interleaved during training.

\begin{algorithm}[t]
\caption{BRC Training Procedure}
\label{alg:brc_training}
\footnotesize
\begin{algorithmic}[1]
\REQUIRE Base learner $f_\theta$ and RCN $\mathrm{RCN}_\phi$
\REQUIRE Hyperparameters $T_{\mathrm{rcn}}$, $T_{\mathrm{res}}$, $K_{\mathrm{rcn}}$,
$\tau_{\mathrm{rcn}}$, and $\rho$
\FOR{each training iteration $t$}
    \STATE Draw labeled data $\mathcal{D}_{\ell}$ and unlabeled data
    $\mathcal{D}_u$.
    \STATE Run the base SSL forward and selection process to obtain
    $\mathcal{L}_{\mathrm{sup}}$, $\mathcal{L}_{\mathrm{pl}}$, and the
    rejected set $\mathcal{D}_{\mathrm{rej}}$.
    \STATE \textbf{// Construct the backbone objective; hold $\phi$ fixed.}
    \IF{$t\geq T_{\mathrm{res}}$ and $\mathcal{D}_{\mathrm{rej}}\neq\varnothing$}
        \STATE Run the RCN on $\mathcal{D}_{\mathrm{rej}}$ and select its top-$\rho$
        samples by utility score as $\mathcal{D}_{\mathrm{res}}$.
        \STATE Compute $\mathcal{L}_{\mathrm{res}}$ by
        Eq.~\eqref{eq:rescue_loss}.
    \ELSE
        \STATE Set $\mathcal{D}_{\mathrm{res}}=\varnothing$ and
        $\mathcal{L}_{\mathrm{res}}=0$.
    \ENDIF
    \STATE Construct $\mathcal{L}_{\mathrm{BRC}}$ by
    Eq.~\eqref{eq:main_loss}.
    \STATE \textbf{// Scheduled RCN update; keep real $\theta$ unchanged.}
    \IF{$t\geq T_{\mathrm{rcn}}$ and $t\bmod K_{\mathrm{rcn}}=0$}
        \STATE Split $\mathcal{D}_u$ into $\mathcal{D}_{\mathrm{low}}$ and
        $\mathcal{D}_{\mathrm{high}}$ using $\tau_{\mathrm{rcn}}$.
        \IF{$\mathcal{D}_{\mathrm{low}}\neq\varnothing$}
            \STATE Run the RCN on $\mathcal{D}_{\mathrm{low}}$.
            \STATE Compute the virtual parameters $\theta'$ by
            Eq.~\eqref{eq:virtual_update}.
            \STATE Compute $\mathcal{L}_{\mathrm{corr}}$ on the labeled and
            high-confidence data by Eq.~\eqref{eq:correction_loss}.
            \STATE Derive $t_j$ and $\mathcal{L}_{\mathrm{score}}$ by
            Eqs.~\eqref{eq:target}--\eqref{eq:score_loss}.
            \STATE Update $\phi$ using $\mathcal{L}_{\mathrm{RCN}}$.
        \ENDIF
    \ENDIF
    \STATE Update $\theta$ using the constructed
    $\mathcal{L}_{\mathrm{BRC}}$.
\ENDFOR
\end{algorithmic}
\end{algorithm}

% ===========================================================
\section{Experiments}
\label{sec:experiments}

% Place the main comparison early so the double-column float appears at the
% top of the page containing the Main Results discussion.
\begin{table*}[t]
\caption{Main results in Top-1 error rate (\%; mean $\pm$ standard deviation).
Lower is better. BRC reduces error in all 30 method--setting pairs,
demonstrating broad applicability across datasets and base learners.}
\label{tab:brc_results}
\centering
\scriptsize
\setlength{\tabcolsep}{1.8pt}
\renewcommand{\arraystretch}{1.02}
\resizebox{\textwidth}{!}{%
\begin{tabular}{lccccccccccc}
\toprule
\multirow{2}{*}{Dataset (labeled samples)} &
\multicolumn{2}{c}{FixMatch} &
\multicolumn{2}{c}{FlexMatch} &
\multicolumn{2}{c}{FreeMatch} &
\multicolumn{2}{c}{OTMatch} &
\multicolumn{2}{c}{RegMixMatch} &
\multirow{2}{*}{Avg. reduction} \\
\cmidrule(lr){2-3}\cmidrule(lr){4-5}\cmidrule(lr){6-7}\cmidrule(lr){8-9}\cmidrule(lr){10-11}
& {Base} & {+BRC} & {Base} & {+BRC} & {Base} & {+BRC} & {Base} & {+BRC} & {Base} & {+BRC} & {$\Delta_{\mathrm{err}}$} \\
\midrule
CIFAR-100 (100)   & 53.19$\pm$1.36 & \textbf{52.74$\boldsymbol{\pm}$0.35} & 42.68$\pm$2.84 & \textbf{38.50$\boldsymbol{\pm}$4.45} & 38.08$\pm$4.06 & \textbf{34.94$\boldsymbol{\pm}$2.04} & 41.48$\pm$2.56 & \textbf{36.51$\boldsymbol{\pm}$3.24} & 39.55$\pm$5.03 & \textbf{38.03$\boldsymbol{\pm}$4.93} & 2.85 \\
CIFAR-100 (200)   & 31.55$\pm$3.86 & \textbf{30.64$\boldsymbol{\pm}$2.23} & 27.08$\pm$1.50 & \textbf{23.18$\boldsymbol{\pm}$1.84} & 21.44$\pm$3.56 & \textbf{20.42$\boldsymbol{\pm}$1.81} & 21.56$\pm$2.70 & \textbf{20.48$\boldsymbol{\pm}$1.96} & 22.18$\pm$5.04 & \textbf{20.99$\boldsymbol{\pm}$4.73} & 1.62 \\
STL-10 (40)       & 18.82$\pm$4.00 & \textbf{14.51$\boldsymbol{\pm}$6.85} & 14.14$\pm$5.54 & \textbf{13.80$\boldsymbol{\pm}$3.28} & 12.78$\pm$3.01 & \textbf{10.55$\boldsymbol{\pm}$2.38} & 13.26$\pm$2.59 & \textbf{12.66$\boldsymbol{\pm}$4.49} & 11.53$\pm$4.40 & \textbf{8.13$\boldsymbol{\pm}$0.66} & 2.18 \\
STL-10 (100)      & 6.01$\pm$1.04 & \textbf{4.94$\boldsymbol{\pm}$0.67} & 6.55$\pm$1.26 & \textbf{6.20$\boldsymbol{\pm}$0.99} & 6.18$\pm$0.76 & \textbf{6.15$\boldsymbol{\pm}$0.88} & 6.36$\pm$0.87 & \textbf{6.19$\boldsymbol{\pm}$0.99} & 6.65$\pm$1.37 & \textbf{6.37$\boldsymbol{\pm}$0.83} & 0.38 \\
TissueMNIST (100) & 55.14$\pm$1.59 & \textbf{54.19$\boldsymbol{\pm}$1.92} & 57.84$\pm$1.81 & \textbf{56.57$\boldsymbol{\pm}$2.54} & 57.38$\pm$2.74 & \textbf{55.66$\boldsymbol{\pm}$1.59} & 57.02$\pm$2.71 & \textbf{55.65$\boldsymbol{\pm}$1.38} & 56.68$\pm$1.27 & \textbf{55.10$\boldsymbol{\pm}$0.61} & 1.38 \\
TissueMNIST (200) & 53.20$\pm$1.21 & \textbf{52.88$\boldsymbol{\pm}$2.19} & 54.49$\pm$2.75 & \textbf{53.97$\boldsymbol{\pm}$2.07} & 53.83$\pm$3.06 & \textbf{53.81$\boldsymbol{\pm}$2.37} & 54.02$\pm$2.67 & \textbf{52.93$\boldsymbol{\pm}$2.23} & 52.54$\pm$1.04 & \textbf{51.95$\boldsymbol{\pm}$2.16} & 0.51 \\
\midrule
Avg. reduction & \multicolumn{2}{c}{1.34} & \multicolumn{2}{c}{1.76} & \multicolumn{2}{c}{1.36} & \multicolumn{2}{c}{1.55} & \multicolumn{2}{c}{1.43} & 1.49 \\
\bottomrule
\end{tabular}
}
\end{table*}

\subsection{Implementation details}
\label{subsec:setup}

We follow the USB protocol for the base training setup, including the
dataset-specific backbone, input resolution, pretrained initialization, and
layer-wise learning-rate decay~\cite{wang2022usb}. We train each model for
$2.048\times10^5$ iterations. For each dataset and base learner, the baseline
and its BRC-augmented counterpart use the same split, backbone, and training
configuration.
For all experimental configurations under each base learner, we use the same
BRC hyperparameters. Supplementary Secs.~1--2 give training and implementation details.

The shared RCN trunk has three 256-dimensional linear layers, each followed by
LeakyReLU, batch normalization, and dropout with probability 0.5. Each output
head is a two-layer MLP. The inner learning rate $\alpha$ follows the current
main-optimizer learning rate, and the virtual update uses second-order
automatic differentiation. To support reproducibility, our implementation
will be made publicly available.

\vspace{-0.45\baselineskip}
\subsection{Datasets and metrics}
\label{subsec:datasets_metrics}

We use six experimental settings:
CIFAR-100 with 100 and 200 labels~\cite{krizhevsky2009cifar}, STL-10 with 40
and 100 labels~\cite{coates2011stl10}, and TissueMNIST with 100 and 200
labels~\cite{medmnist2021}. We report Top-1 error rate as the mean $\pm$
standard deviation, computed over three random seeds.

\subsection{Baselines}
\label{subsec:baselines}

We evaluate BRC as an add-on to five representative confidence-thresholded SSL
methods. FixMatch~\cite{sohn2020fixmatch} uses a fixed
confidence threshold. FlexMatch~\cite{zhang2021flexmatch} and
FreeMatch~\cite{wang2023freematch} use adaptive thresholding strategies.
OTMatch~\cite{tan2024otmatch} and RegMixMatch~\cite{han2025regmixmatch} are
recent strong baselines with additional regularization or matching objectives.
This set covers fixed, adaptive, and recent high-performing confidence-thresholded
SSL methods. In each dataset-label setting, we compare the original base learner
with its BRC-augmented counterpart.

\subsection{Main results}
\label{subsec:results}

Table~\ref{tab:brc_results} shows that BRC reduces error in all 30
method--setting pairs, with an average reduction of 1.49 percentage points
over six settings and five base learners. The improvements are more pronounced in
low-label regimes. For instance, on CIFAR-100 (100 labels), STL-10 (40 labels),
and TissueMNIST (100 labels), BRC reduces the average error rate of the five
base learners by 2.14 percentage points. This suggests that BRC is particularly
beneficial when labeled supervision is scarce, because the rejected
boundary-sample pool---due to limited initial training---contains more useful
samples that would be discarded without BRC. However, on relatively simple
datasets with sufficient labeled data, such as STL-10 with 100 labels, BRC
shows only marginal improvement (0.38 percentage points). This is
understandable, since with enough labels the base learner already discards
low-confidence samples fairly accurately, so BRC refinement is largely
unnecessary.

\subsection{Ablation study}
\label{subsec:ablation}

We next examine whether BRC's gains require its two core components: learned
selection of rescue candidates and correction of their pseudo-labels.

Base provides the reference without BRC rescue, while Full BRC uses both learned
selection and label correction. To isolate selection, the ``random selection''
variant preserves corrected rescue but chooses candidates randomly. To isolate
correction, the ``w/o correction'' variant retains learned selection but reuses
the original pseudo-labels. We evaluate all variants with FlexMatch, FreeMatch,
and RegMixMatch under the same setting.

\begin{table}[t]
\caption{Ablations of learned selection and label correction. Entries are Top-1
error rate (\%; mean $\pm$ standard deviation); lower is better. Full BRC
performs best; random selection weakens the gains, while removing correction
performs worse than Base.}
\label{tab:ablation}
\centering
\scriptsize
\setlength{\tabcolsep}{2.4pt}
\renewcommand{\arraystretch}{0.96}
\begin{tabular}{@{}lccc@{}}
\toprule
Variant & FlexMatch & FreeMatch & RegMixMatch \\
\midrule
Base & 42.68$\pm$2.84 & 38.08$\pm$4.06 & 39.55$\pm$5.03 \\
Full BRC & \textbf{38.50$\boldsymbol{\pm}$4.45} & \textbf{34.94$\boldsymbol{\pm}$2.04} & \textbf{38.03$\boldsymbol{\pm}$4.93} \\
Random selection & 42.91$\pm$2.58 & 36.59$\pm$1.54 & 40.25$\pm$3.12 \\
w/o correction & 45.21$\pm$4.00 & 38.27$\pm$2.15 & 44.51$\pm$2.21 \\
\bottomrule
\end{tabular}
\end{table}

Table~\ref{tab:ablation} shows that Full BRC consistently achieves the lowest
error rates across all three base learners. Replacing learned selection with
random selection consistently weakens the result, showing that the choice of
rescue candidates matters. Removing correction is even more damaging: it
raises the error above the Base level in every case, demonstrating that directly
reusing rejected samples with their original pseudo-labels can be harmful.
Together, these results show that BRC's gains depend on both selecting
beneficial rejected samples and correcting their targets before reuse.

\begin{figure*}[t]
\centering
\begin{minipage}[t]{0.49\textwidth}
\centering
\setcounter{subfigure}{0}
\begin{subfigure}[t]{0.49\linewidth}
\centering
\includegraphics[width=\linewidth]{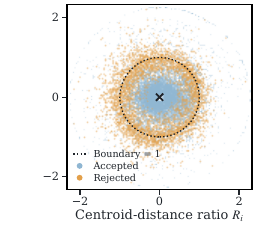}
\caption{Centroid-based boundary map}
\end{subfigure}
\hfill
\begin{subfigure}[t]{0.49\linewidth}
\centering
\includegraphics[width=\linewidth]{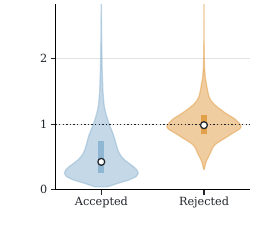}
\caption{Full-sample boundary distribution}
\end{subfigure}
\caption{Distributions of accepted and rejected samples. Rejected samples
concentrate near the decision boundary, showing that confidence thresholding
produces a boundary-biased rejected pool.}
\label{fig:centroid-boundary-analysis}
\end{minipage}
\hfill
\begin{minipage}[t]{0.49\textwidth}
\centering
\setcounter{subfigure}{0}
\begin{subfigure}[t]{0.49\linewidth}
\centering
\includegraphics[width=\linewidth]{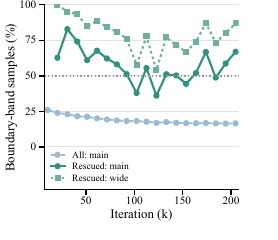}
\caption{Rescue concentration}
\end{subfigure}
\hfill
\begin{subfigure}[t]{0.49\linewidth}
\centering
\includegraphics[width=\linewidth]{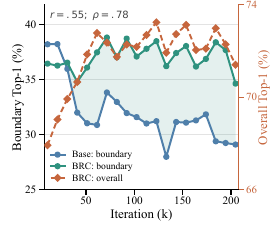}
\caption{Boundary vs. overall Top-1}
\end{subfigure}
\caption{Rescued-sample distribution and boundary accuracy during training.
(a) Boundary-sample prevalence in the full and rescued pools. (b) Boundary
Top-1 of Base and BRC together with BRC overall Top-1. The two panels show that
BRC prioritizes boundary-near samples and improves their predictions.}
\label{fig:rescue-dynamics}
\end{minipage}
\end{figure*}

\subsection{Distribution analysis of rejected samples}
\label{subsec:sample-distribution-analysis}

BRC is built on the central hypothesis that the rejected pool produced by
confidence thresholding is not a random subset of the unlabeled data, but is
systematically biased toward the decision boundary.

To examine this hypothesis, we analyze the midpoint checkpoint of FixMatch on
CIFAR-100 with 200 labels (iteration $102{,}400$ of $204{,}800$). For all
$50{,}000$ training samples, we record their accepted/rejected status and use a
centroid-distance ratio $R_i$ to measure proximity to a class boundary in
feature space. We first extract $L_2$-normalized features and compute an
$L_2$-normalized centroid for each ground-truth class. For each sample, $R_i$
is the cosine distance to its ground-truth class centroid divided by the
distance to the nearest competing-class centroid. Thus, $R_i<1$ indicates that
the sample is closer to its ground-truth centroid, whereas $R_i>1$ indicates
that it is closer to a competing-class centroid. We regard $R_i=1$ as an
approximate class boundary and values near one as boundary-near. Ground-truth
labels are used only to construct the centroids for this analysis and are not
used during training. Calculation details are in Supplementary Sec.~4.1.

Figure~\ref{fig:centroid-boundary-analysis} reveals a clear difference between
the two groups. Rejected samples have mean and median ratios of $1.006$ and
$0.985$, compared with $0.565$ and $0.425$ for accepted samples. Because values
near one correspond to the approximate boundary, these distributional
differences confirm that the rejected pool is not a random subset of the overall
sample population, but is markedly biased toward the decision boundary.
\subsection{Rescue dynamics and boundary refinement}
\label{subsec:boundary-analysis}

Section~\ref{subsec:sample-distribution-analysis} establishes that confidence
thresholding produces a boundary-biased rejected pool. We next examine whether
BRC exploits this structure as intended by focusing its rescue on boundary-near
samples and improving their classification accuracy.

We inspect 20 matched Base/BRC checkpoints of FlexMatch on CIFAR-100 with 200
labels and seed 1, spaced by $10{,}240$ iterations. At each checkpoint, Base
defines strict
($R_i\in[0.90,1.10]$), main ($R_i\in[0.80,1.25]$), and wide
($R_i\in[0.67,1.50]$) boundary bands. We replay BRC selection to compare the
prevalence of boundary samples in the rescued set and the full pool, and we
evaluate Base and BRC on exactly the same Base-defined main-band samples.

Figure~\ref{fig:rescue-dynamics}(a) shows that main-band samples constitute
only $16.45\%$--$25.99\%$ of the pool but $36.12\%$--$82.83\%$ of the rescued
set, corresponding to a $2.09\times$--$4.04\times$ enrichment. Thus, the
selector primarily targets boundary-relevant candidates. In
Figure~\ref{fig:rescue-dynamics}(b), BRC outperforms Base on the same boundary
samples at 18 of 20 checkpoints, with a maximum gain of $9.00$ points.
Boundary and overall Top-1 follow similar trends (Pearson $r=0.55$; Spearman
$\rho=0.78$). Together, the boundary enrichment and consistent accuracy gains
show that BRC behaves as intended: it prioritizes boundary-near samples during
training and improves their predictions.

\subsection{Comparison with lower-threshold SSL}
\label{subsec:fixed-set-control}

Although Figure~\ref{fig:rescue-dynamics} shows that BRC improves boundary
predictions, a simpler explanation is that the gain comes merely from accepting
more low-confidence samples. We therefore introduce a threshold-control
experiment to isolate this possibility.

Because boundary-band membership changes during training, we use an
intermediate Base checkpoint to define the strict, main, and wide boundary sets
once and retain the same sample IDs when evaluating every model at its best
checkpoint. Besides Base and BRC, we conduct several control experiments using
lower FlexMatch thresholds. As these settings exhibit similar trends, we
present the $0.90$ threshold as a representative control. Relative to the
default threshold of $0.95$, this control expands the acceptance range but
performs neither learned selection nor label correction. We compare Top-1 accuracy and negative log-likelihood (NLL).

\begin{figure}[!t]
\centering
\begin{subfigure}{\columnwidth}
\centering
\includegraphics[width=\linewidth]{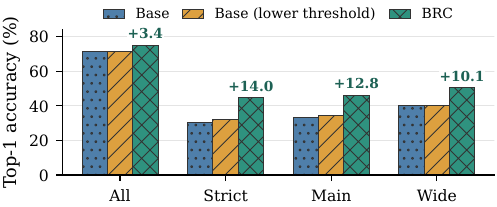}
\caption{Top-1 accuracy}
\end{subfigure}
\vspace{-0.6ex}
\begin{subfigure}{\columnwidth}
\centering
\includegraphics[width=\linewidth]{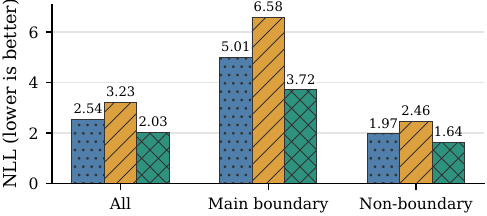}
\caption{Negative log-likelihood}
\end{subfigure}
\caption{Top-1 accuracy (a) and NLL (b) of Base, a lower-threshold control, and
BRC on fixed full, non-boundary, and boundary sample sets. BRC substantially
improves boundary predictions, whereas lowering the threshold does not,
supporting the learned select--correct--reuse mechanism.}
\label{fig:fixed-set-control}
\end{figure}

Figure~\ref{fig:fixed-set-control} shows that simply expanding acceptance does
not reproduce BRC's gains. In panel (a), lowering the threshold changes
main-boundary Top-1 only from $33.28\%$ to $34.15\%$ and slightly reduces
overall Top-1, whereas BRC raises main-boundary Top-1 to $46.08\%$---$12.80$
points above Base and $11.93$ points above the threshold control. Panel (b)
shows the same contrast in predictive quality: on the main-boundary set,
lowering the threshold worsens NLL from $5.01$ to $6.58$, whereas BRC reduces it
to $3.72$. The full and non-boundary sets exhibit the same qualitative pattern.
The threshold control therefore fails to reproduce either the accuracy or NLL
improvement of BRC. This contrast rules out relaxed acceptance as a sufficient
explanation and supports learned selection and label correction as the key
source of BRC's gains.

\subsection{BRC hyperparameter discussion}
\label{subsec:hyperparameters}

We further assess whether BRC remains effective under reasonable
hyperparameter variations rather than depending on a narrowly tuned
configuration.

Using FlexMatch on CIFAR-100 with 200 labels, we vary the RCN confidence
threshold $\tau_{\mathrm{rcn}}$, rescue ratio $\rho$, and rescue start
iteration $T_{\mathrm{res}}$ one at a time while keeping the remaining settings
fixed.

\begin{figure}[t]
\centering
\includegraphics[width=\columnwidth]{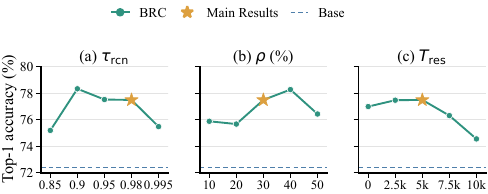}
\caption{Hyperparameter sensitivity of BRC. The star marks the setting used for
the main results, and the dashed line shows the Base accuracy. BRC outperforms
Base under every tested configuration, indicating robustness to reasonable
hyperparameter choices.}
\label{fig:brc-hparam-sensitivity}
\end{figure}

Figure~\ref{fig:brc-hparam-sensitivity} shows that every tested configuration
outperforms Base, demonstrating that the improvement is not tied to a single
hyperparameter choice within the tested range. Performance remains strong for
$\tau_{\mathrm{rcn}}=0.90$--$0.98$, $\rho=30\%$--$40\%$, and
$T_{\mathrm{res}}\leq5{,}000$. The decline under excessively delayed rescue
also supports the intended mechanism: useful rejected samples cannot
contribute while rescue remains inactive. The values used for the main results
lie within the stable region and are shared across all FlexMatch settings.

\subsection{Complexity analysis}
\label{subsec:complexity}

We next assess whether BRC provides a practical trade-off between predictive
performance and computational overhead.

Using the CIFAR-100 FlexMatch setting, we measure error rate, trainable
parameters, and wall-clock training time while varying the RCN update interval
$K_{\mathrm{rcn}}$. A larger interval invokes the RCN update less frequently,
thereby reducing its training-time overhead.

\begin{table}[t]
\caption{Error rate, trainable parameters, and training time under different RCN
update intervals. The RCN adds only 1.9\% trainable parameters, while its update
interval provides a controllable accuracy--efficiency trade-off.}
\label{tab:complexity}
\centering
\scriptsize
\setlength{\tabcolsep}{1.8pt}
\renewcommand{\arraystretch}{0.98}
\begin{tabular}{@{}lccccc@{}}
\toprule
Method & $K_{\mathrm{rcn}}$ & Err. & Params & Wall time & Time ratio \\
& & (\%) & (M) & (h) & ($\times$) \\
\midrule
FlexMatch & \NA & 27.59 & 21.44 & 9.39 & 1.00 \\
BRC & 2  & 23.70 & 21.85 & 17.63 & 1.88 \\
BRC & 5  & 20.34 & 21.85 & 13.33 & 1.42 \\
BRC & 10 & 22.50 & 21.85 & 11.48 & 1.22 \\
BRC & 20 & 22.19 & 21.85 & 10.04 & 1.07 \\
\bottomrule
\end{tabular}
\end{table}

Table~\ref{tab:complexity} shows that the RCN adds only 0.415M parameters to the
21.44M-parameter backbone, an increase of 1.9\%, and is removed at inference so
that prediction cost is unchanged. The update interval provides a controllable
accuracy--efficiency trade-off: $K_{\mathrm{rcn}}=5$ reduces the error from
$27.59\%$ to $20.34\%$ at $1.42\times$ the baseline training time, whereas
$K_{\mathrm{rcn}}=20$ still achieves $22.19\%$ error with only
$1.07\times$ training time. These results show that BRC can retain substantial
performance gains with modest parameter and training-time overhead.

\subsection{Qualitative case study}
\label{subsec:case-study}

\begin{figure}[t]
\centering
\includegraphics[width=0.96\columnwidth]{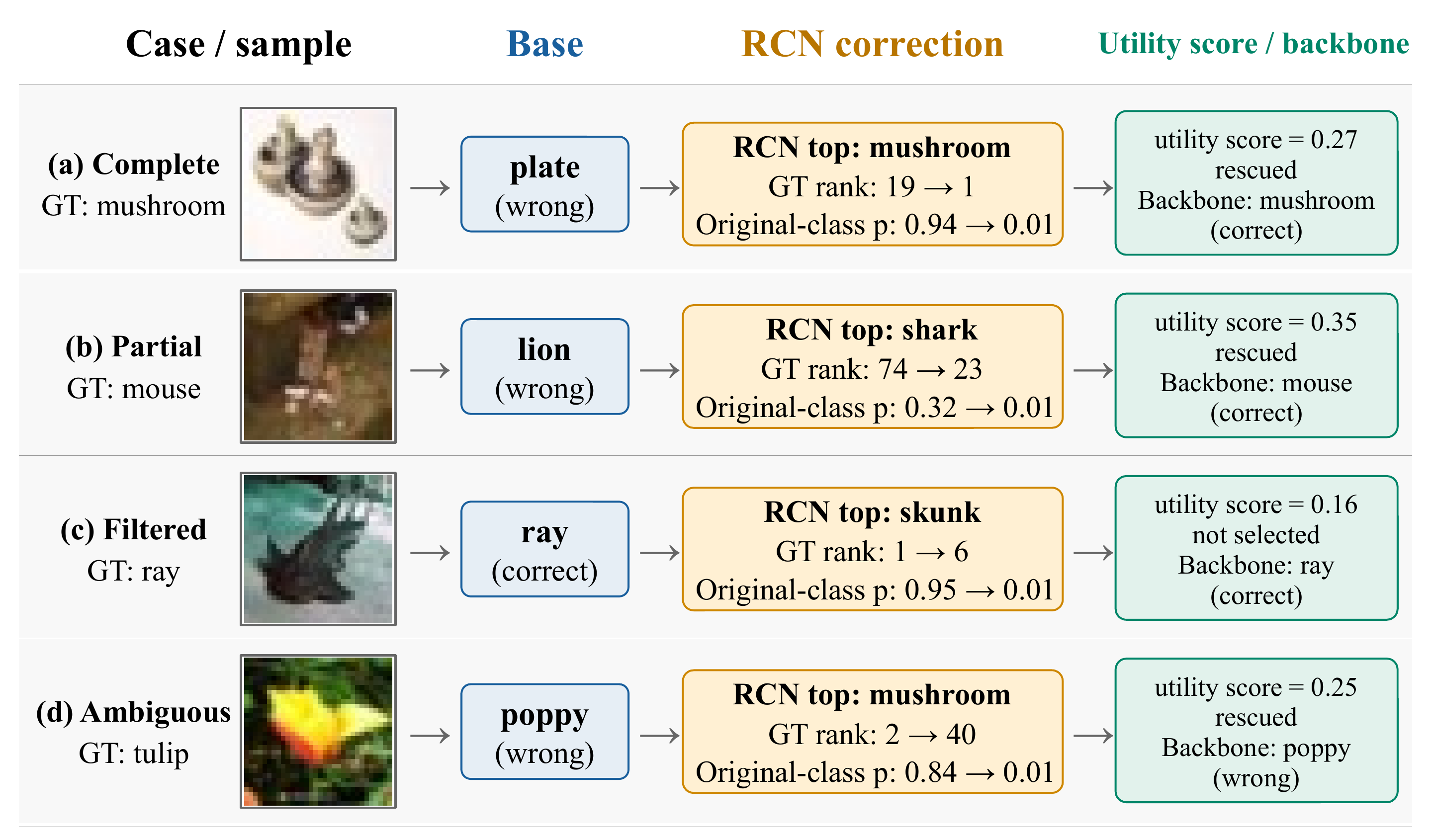}
\caption{Sample-level BRC cases. GT rank is the ground-truth class rank (1 is
best); Original-class p is the probability assigned to the original
pseudo-label class before and after correction. The
panels show complete correction, partial improvement, correct filtering, and a
failure case; the final column reports the utility score, rescue decision, and
the BRC-trained backbone prediction.}
\label{fig:case-study}
\end{figure}

To make BRC's sample-level behavior explicit, we inspect representative
rejected samples and visualize how prediction, correction, and utility scores interact. This
post-hoc analysis reports the Base prediction, RCN-corrected target, rescue
decision based on the utility score, and final prediction of the BRC-trained backbone. The orange column
shows the RCN correction, whereas ``Backbone'' in the final column denotes the
backbone prediction after BRC training. In Figure~\ref{fig:case-study}, GT rank
denotes the position of the ground-truth class after sorting class probabilities
in descending order, where rank 1 is best. Original-class p denotes the
probability assigned to the original pseudo-label class before and after RCN
correction.

Figure~\ref{fig:case-study}(a) shows complete correction: the RCN moves the
ground-truth class from rank 19 to rank 1 and suppresses the original wrong
class; the final backbone predicts mushroom correctly. Figure~\ref{fig:case-study}(b) shows
partial soft correction of an initially incorrect Base prediction: the
ground-truth rank improves from 74 to 23 and the original-class probability
drops from $0.32$ to $0.01$. Although the ground truth does not become the RCN's
Top-1 class, the corrected soft target is substantially less misleading, and
the final backbone correctly predicts mouse.
Figure~\ref{fig:case-study}(c) shows filtering based on utility scores: the proposed
correction would demote the ground truth from rank 1 to rank 6, so its low
utility score leaves the sample unselected for rescue and prevents the harmful
corrected target from entering the unsupervised loss; the final backbone
prediction remains ray. Figure~\ref{fig:case-study}(d) shows a failure mode: visual ambiguity between
tulip and poppy at $32\times32$ resolution may mislead correction. The RCN
demotes GT rank from 2 to 40, yet a utility score of $0.25$ still leads to
rescue, and the final backbone predicts poppy. Future work could improve class
discrimination and introduce a minimum utility-score requirement. Supplementary Sec.~6.1 shows four additional selected/rescued examples.

\subsection{Generalization to semantic segmentation}
\label{subsec:segmentation-transfer}

To assess whether BRC's \emph{select--correct--reuse} principle extends beyond
image classification, we explore semi-supervised semantic segmentation.
Boundary pixels are difficult to classify and are frequently rejected by
confidence filtering, making segmentation a natural testbed for evaluating BRC
at the pixel level~\cite{wang2022u2pl,liu2025confidence}.

PixelBRC treats each pixel as a training sample. The base segmentation method
first divides unlabeled pixels into accepted and rejected sets using its
original confidence rule. PixelBRC then predicts corrected labels and utility scores for
rejected pixels, selects the most useful candidates, and returns their corrected
soft labels to segmentation training. While preserving BRC's core
\emph{select--correct--reuse} principle, PixelBRC adapts the optimization of its
two heads to dense prediction: both the correction and score heads are trained
directly from labeled pixel-level supervision rather than through the
meta-learning objective used for classification. Supplementary Sec.~3 gives PixelBRC implementation details.

\begin{figure}[t]
\centering
\includegraphics[width=0.96\columnwidth]{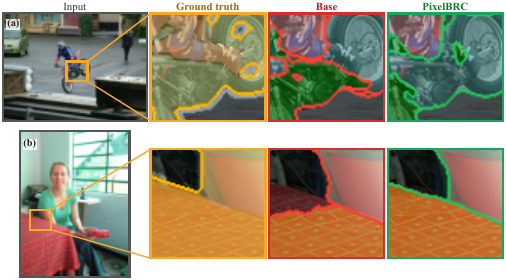}
\caption{Pixel-level case studies. In each row, yellow highlights the
ground-truth region, while red and green contours denote the Base and PixelBRC
boundaries. PixelBRC follows the ground truth more closely in both cases.}
\label{fig:pixelbrc-case}
\end{figure}

We evaluate PixelBRC with representative semi-supervised segmentation methods~\cite{yang2025unimatchv2}
and datasets~\cite{brostow2009camvid,everingham2010pascal,zhou2017ade20k,
gould2009decomposing}. As shown in Table~\ref{tab:segmentation-transfer}, PixelBRC
improves every evaluated setting by $1.03$--$2.39$ mIoU. Despite this
task-specific change in head supervision, the consistent gains provide evidence
that BRC's core \emph{select--correct--reuse} principle generalizes from image
classification to pixel-level dense prediction. Figure~\ref{fig:pixelbrc-case}
shows two local structures where PixelBRC corrects Base boundary errors,
making the aggregate improvement spatially interpretable. Supplementary Sec.~6.2 shows six additional segmentation cases.

\begin{table}[t]
\caption{Exploratory transfer to semi-supervised semantic segmentation. Entries
are best validation mIoU (\%); higher is better. PixelBRC improves all evaluated
settings by $1.03$--$2.39$ mIoU, supporting BRC's cross-task generalization to
dense prediction.}
\label{tab:segmentation-transfer}
\centering
\scriptsize
\setlength{\tabcolsep}{2.5pt}
\renewcommand{\arraystretch}{0.94}
\begin{tabular}{@{}llccc@{}}
\toprule
Dataset (split) & Base learner & Base & +PixelBRC & $\Delta$ \\
\midrule
CamVid (1/4) & Dense FreeMatch & 79.91 & \textbf{82.30} & +2.39 \\
PASCAL VOC (1/8) & FixMatch & 85.94 & \textbf{87.55} & +1.61 \\
ADE20K (1/16) & UniMatch V2 & 45.34 & \textbf{46.75} & +1.41 \\
Stanford BG (1/4) & Teacher--student & 76.50 & \textbf{77.67} & +1.17 \\
CamVid (1/4) & FixMatch & 82.04 & \textbf{83.07} & +1.03 \\
\bottomrule
\end{tabular}
\end{table}

% ===========================================================
\section{Conclusion}
\label{sec:conclusion}

We presented BRC, an add-on for confidence-thresholded SSL that recovers
useful supervision from discarded samples. Our analysis shows that confidence filtering produces a
rejected pool concentrated near class boundaries, motivating selective rescue
as a post-selection problem. BRC preserves the base method's selection rule,
estimates the training utility of discarded candidates, and returns selected
samples to the unsupervised loss with corrected soft pseudo-labels. Across five
base learners and six classification settings, BRC consistently reduces error,
achieving an average reduction of 1.49 percentage points. Component ablations
support the importance of both learned selection and label correction. Boundary
analyses further show that BRC preferentially rescues boundary-near samples and
uses their corrected pseudo-labels for training, thereby improving the backbone's
accuracy on a fixed set of boundary samples by 12.80 percentage points. An
exploratory extension to semantic segmentation yields improvements of
1.03--2.39 mIoU, providing initial evidence that the select--correct--reuse
principle extends to dense prediction. These findings highlight the value of
selectively recovering boundary supervision discarded by confidence filtering.

% ===========================================================
{
\small
\bibliographystyle{ieeenat_fullname}
\bibliography{main}

\begin{thebibliography}{47}
\providecommand{\natexlab}[1]{#1}
\providecommand{\url}[1]{\texttt{#1}}
\expandafter\ifx\csname urlstyle\endcsname\relax
  \providecommand{\doi}[1]{doi: #1}\else
  \providecommand{\doi}{doi: \begingroup \urlstyle{rm}\Url}\fi

\bibitem[Arazo et~al.(2020)Arazo, Ortego, Albert, O'Connor, and
  McGuinness]{arazo2020pseudo}
Eric Arazo, Diego Ortego, Paul Albert, Noel~E. O'Connor, and Kevin McGuinness.
\newblock Pseudo-labeling and confirmation bias in deep semi-supervised
  learning.
\newblock In \emph{Proc. Int. Joint Conf. Neural Netw. (IJCNN)}, pages 1--8,
  2020.

\bibitem[Berthelot et~al.(2019)Berthelot, Carlini, Goodfellow, Papernot,
  Oliver, and Raffel]{berthelot2019mixmatch}
David Berthelot, Nicholas Carlini, Ian Goodfellow, Nicolas Papernot, Avital
  Oliver, and Colin Raffel.
\newblock {MixMatch}: A holistic approach to semi-supervised learning.
\newblock In \emph{Adv. Neural Inf. Process. Syst. (NeurIPS)}, 2019.

\bibitem[Berthelot et~al.(2020)Berthelot, Carlini, Cubuk, Kurakin, Sohn, Zhang,
  and Raffel]{berthelot2020remixmatch}
David Berthelot, Nicholas Carlini, Ekin~D. Cubuk, Alex Kurakin, Kihyuk Sohn,
  Han Zhang, and Colin Raffel.
\newblock {ReMixMatch}: Semi-supervised learning with distribution alignment
  and augmentation anchoring.
\newblock In \emph{Int. Conf. Learn. Represent. (ICLR)}, 2020.

\bibitem[Brostow et~al.(2009)Brostow, Fauqueur, and Cipolla]{brostow2009camvid}
Gabriel~J. Brostow, Julien Fauqueur, and Roberto Cipolla.
\newblock Semantic object classes in video: A high-definition ground truth
  database.
\newblock \emph{Pattern Recognition Letters}, 30\penalty0 (2):\penalty0 88--97,
  2009.

\bibitem[Chen et~al.(2023)Chen, Tao, Fan, Wang, Wang, Schiele, Xie, Raj, and
  Savvides]{chen2023softmatch}
Hao Chen, Ran Tao, Yue Fan, Yidong Wang, Jindong Wang, Bernt Schiele, Xing Xie,
  Bhiksha Raj, and Marios Savvides.
\newblock {SoftMatch}: Addressing the quantity-quality trade-off in
  semi-supervised learning.
\newblock In \emph{Int. Conf. Learn. Represent. (ICLR)}, 2023.

\bibitem[Cheng et~al.(2025)Cheng, Lu, Tian, Zhao, Chang, and
  Du]{cheng2025cgmatch}
Bo Cheng, Jueqing Lu, Yuan Tian, Haifeng Zhao, Yi Chang, and Lan Du.
\newblock {CGMatch}: A different perspective of semi-supervised learning.
\newblock In \emph{Proc. IEEE/CVF Conf. Comput. Vis. Pattern Recognit. (CVPR)},
  pages 15381--15391, 2025.

\bibitem[Coates et~al.(2011)Coates, Ng, and Lee]{coates2011stl10}
Adam Coates, Andrew Ng, and Honglak Lee.
\newblock An analysis of single-layer networks in unsupervised feature
  learning.
\newblock In \emph{Proc. Int. Conf. Artif. Intell. Statist. (AISTATS)}, pages
  215--223, 2011.

\bibitem[Cubuk et~al.(2020)Cubuk, Zoph, Shlens, and Le]{cubuk2020randaugment}
Ekin~D. Cubuk, Barret Zoph, Jonathon Shlens, and Quoc~V. Le.
\newblock {RandAugment}: Practical automated data augmentation with a reduced
  search space.
\newblock In \emph{Proc. IEEE/CVF Conf. Comput. Vis. Pattern Recognit.
  Workshops (CVPRW)}, pages 702--703, 2020.

\bibitem[Everingham et~al.(2010)Everingham, {Van Gool}, Williams, Winn, and
  Zisserman]{everingham2010pascal}
Mark Everingham, Luc {Van Gool}, Christopher K.~I. Williams, John Winn, and
  Andrew Zisserman.
\newblock The {PASCAL} visual object classes ({VOC}) challenge.
\newblock \emph{International Journal of Computer Vision}, 88\penalty0
  (2):\penalty0 303--338, 2010.

\bibitem[Finn et~al.(2017)Finn, Abbeel, and Levine]{finn2017maml}
Chelsea Finn, Pieter Abbeel, and Sergey Levine.
\newblock Model-agnostic meta-learning for fast adaptation of deep networks.
\newblock In \emph{Proc. Int. Conf. Mach. Learn. (ICML)}, pages 1126--1135,
  2017.

\bibitem[Gould et~al.(2009)Gould, Fulton, and Koller]{gould2009decomposing}
Stephen Gould, Richard Fulton, and Daphne Koller.
\newblock Decomposing a scene into geometric and semantically consistent
  regions.
\newblock In \emph{Proc. IEEE Int. Conf. Comput. Vis. (ICCV)}, pages 1--8,
  2009.

\bibitem[Han et~al.(2025)Han, Yuan, Wei, and Yu]{han2025regmixmatch}
Haorong Han, Jidong Yuan, Chixuan Wei, and Zhongyang Yu.
\newblock {RegMixMatch}: Optimizing mixup utilization in semi-supervised
  learning.
\newblock In \emph{Proc. AAAI Conf. Artif. Intell.}, pages 17032--17040, 2025.

\bibitem[Han and Jiang(2026)]{han2026paf}
Yinan Han and Qing-Yuan Jiang.
\newblock {PAF}: Perturbation-aware filtering for open-set semi-supervised
  learning.
\newblock In \emph{Proc. IEEE/CVF Conf. Comput. Vis. Pattern Recognit. (CVPR)},
  pages 24803--24812, 2026.

\bibitem[Hariharan et~al.(2011)Hariharan, Arbelaez, Bourdev, Maji, and
  Malik]{hariharan2011semantic}
Bharath Hariharan, Pablo Arbelaez, Lubomir Bourdev, Subhransu Maji, and
  Jitendra Malik.
\newblock Semantic contours from inverse detectors.
\newblock In \emph{Proc. IEEE Int. Conf. Comput. Vis. (ICCV)}, pages 991--998,
  2011.

\bibitem[He et~al.(2022)He, Chen, Xie, Li, Doll{\'a}r, and Girshick]{he2022mae}
Kaiming He, Xinlei Chen, Saining Xie, Yanghao Li, Piotr Doll{\'a}r, and Ross
  Girshick.
\newblock Masked autoencoders are scalable vision learners.
\newblock In \emph{Proc. IEEE/CVF Conf. Comput. Vis. Pattern Recognit. (CVPR)},
  pages 16000--16009, 2022.

\bibitem[Iwata et~al.(2025)Iwata, Kumagai, and Ida]{iwata2025regularization}
Tomoharu Iwata, Atsutoshi Kumagai, and Yasutoshi Ida.
\newblock Meta-learning task-specific regularization weights for few-shot
  linear regression.
\newblock In \emph{Proc. Int. Conf. Artif. Intell. Stat. (AISTATS)}, pages
  3052--3060, 2025.

\bibitem[Krizhevsky(2009)]{krizhevsky2009cifar}
Alex Krizhevsky.
\newblock Learning multiple layers of features from tiny images.
\newblock Technical report, University of Toronto, 2009.

\bibitem[Lee(2013)]{lee2013pseudo}
Dong-Hyun Lee.
\newblock {Pseudo-Label}: The simple and efficient semi-supervised learning
  method for deep neural networks.
\newblock In \emph{Workshop on Challenges in Representation Learning, ICML},
  2013.

\bibitem[Liu et~al.(2026)Liu, He, and Liu]{liu2026covar}
Jinshi Liu, Lei He, and Pan Liu.
\newblock {CoVar}: Confidence-variance-guided pseudo-label selection for
  semi-supervised learning.
\newblock \emph{arXiv preprint arXiv:2601.11670}, 2026.
\newblock Version 3.

\bibitem[Liu and Liu(2025)]{liu2025confidence}
Pan Liu and Jinshi Liu.
\newblock When confidence fails: Revisiting pseudo-label selection in
  semi-supervised semantic segmentation.
\newblock In \emph{Proc. IEEE/CVF Int. Conf. Comput. Vis. (ICCV)}, pages
  21874--21884, 2025.

\bibitem[Mai et~al.(2026)Mai, Sun, Chen, Li, Wang, Li, He, and
  Wang]{mai2026hess}
Huayu Mai, Rui Sun, Yujia Chen, Wangkai Li, Bingzhou Wang, Aibing Li, Zhangyu
  He, and Yuan Wang.
\newblock From {Softmax} to {Dirichlet}: Evidential learning for
  semi-supervised semantic segmentation.
\newblock In \emph{Proc. IEEE/CVF Conf. Comput. Vis. Pattern Recognit. (CVPR)},
  pages 27697--27707, 2026.

\bibitem[Nichol et~al.(2018)Nichol, Achiam, and Schulman]{nichol2018firstorder}
Alex Nichol, Joshua Achiam, and John Schulman.
\newblock On first-order meta-learning algorithms.
\newblock \emph{arXiv preprint arXiv:1803.02999}, 2018.

\bibitem[Oquab et~al.(2024)Oquab, Darcet, Moutakanni, Vo, Szafraniec, Khalidov,
  Fernandez, Haziza, Massa, El-Nouby, Assran, Ballas, Galuba, Howes, Huang, Li,
  Misra, Rabbat, Sharma, Synnaeve, Xu, J{\'e}gou, Mairal, Labatut, Joulin, and
  Bojanowski]{oquab2024dinov2}
Maxime Oquab, Timoth{\'e}e Darcet, Th{\'e}o Moutakanni, Huy~V. Vo, Marc
  Szafraniec, Vasil Khalidov, Pierre Fernandez, Daniel Haziza, Francisco Massa,
  Alaaeldin El-Nouby, Mahmoud Assran, Nicolas Ballas, Wojciech Galuba, Russell
  Howes, Po-Yao Huang, Shang-Wen Li, Ishan Misra, Michael Rabbat, Vasu Sharma,
  Gabriel Synnaeve, Hu Xu, Herv{\'e} J{\'e}gou, Julien Mairal, Patrick Labatut,
  Armand Joulin, and Piotr Bojanowski.
\newblock {DINOv2}: Learning robust visual features without supervision.
\newblock \emph{Transactions on Machine Learning Research}, 2024.

\bibitem[Pham et~al.(2021)Pham, Dai, Xie, and Le]{pham2021meta}
Hieu Pham, Zihang Dai, Qizhe Xie, and Quoc~V. Le.
\newblock Meta pseudo labels.
\newblock In \emph{Proc. IEEE/CVF Conf. Comput. Vis. Pattern Recognit. (CVPR)},
  pages 11557--11568, 2021.

\bibitem[Ranftl et~al.(2021)Ranftl, Bochkovskiy, and Koltun]{ranftl2021dpt}
Ren{\'e} Ranftl, Alexey Bochkovskiy, and Vladlen Koltun.
\newblock Vision transformers for dense prediction.
\newblock In \emph{Proc. IEEE/CVF Int. Conf. Comput. Vis. (ICCV)}, pages
  12179--12188, 2021.

\bibitem[Sohn et~al.(2020)Sohn, Berthelot, Li, Zhang, Carlini, Cubuk, Kurakin,
  Zhang, and Raffel]{sohn2020fixmatch}
Kihyuk Sohn, David Berthelot, Chun-Liang Li, Zizhao Zhang, Nicholas Carlini,
  Ekin~D. Cubuk, Alex Kurakin, Han Zhang, and Colin Raffel.
\newblock {FixMatch}: Simplifying semi-supervised learning with consistency and
  confidence.
\newblock In \emph{Adv. Neural Inf. Process. Syst. (NeurIPS)}, 2020.

\bibitem[Sosea and Caragea(2023)]{sosea2023marginmatch}
Tiberiu Sosea and Cornelia Caragea.
\newblock {MarginMatch}: Improving semi-supervised learning with
  pseudo-margins.
\newblock In \emph{Proc. IEEE/CVF Conf. Comput. Vis. Pattern Recognit. (CVPR)},
  pages 15773--15782, 2023.

\bibitem[Sun et~al.(2024)Sun, Yang, Zhang, Cheng, and Hou]{sun2024corrmatch}
Boyuan Sun, Yuqi Yang, Le Zhang, Ming-Ming Cheng, and Qibin Hou.
\newblock {CorrMatch}: Label propagation via correlation matching for
  semi-supervised semantic segmentation.
\newblock In \emph{Proc. IEEE/CVF Conf. Comput. Vis. Pattern Recognit. (CVPR)},
  pages 3097--3107, 2024.

\bibitem[Tan et~al.(2024)Tan, Zheng, and Huang]{tan2024otmatch}
Zhiquan Tan, Kaipeng Zheng, and Weiran Huang.
\newblock {OTMatch}: Improving semi-supervised learning with optimal transport.
\newblock In \emph{Proc. Int. Conf. Mach. Learn. (ICML)}, pages 47667--47680,
  2024.

\bibitem[Vishwakarma et~al.(2025)Vishwakarma, Chen, Namburi~Gnvv, Tay, Vinayak,
  and Sala]{vishwakarma2025rethinking}
Harit Vishwakarma, Yi Chen, Satya Sai~Srinath Namburi~Gnvv, Sui~Jiet Tay,
  Ramya~Korlakai Vinayak, and Frederic Sala.
\newblock Rethinking confidence scores and thresholds in pseudolabeling-based
  {SSL}.
\newblock In \emph{Proc. Int. Conf. Mach. Learn. (ICML)}, pages 61582--61600,
  2025.

\bibitem[Wang and Arora(2024)]{wang2024metastability}
Yunjuan Wang and Raman Arora.
\newblock On the stability and generalization of meta-learning.
\newblock In \emph{Adv. Neural Inf. Process. Syst. (NeurIPS)}, 2024.

\bibitem[Wang et~al.(2022{\natexlab{a}})Wang, Chen, Fan, Sun, Tao, Hou, Wang,
  Yang, Zhou, Guo, Qi, Wu, Li, Nakamura, Ye, Savvides, Raj, Shinozaki, Schiele,
  Wang, Xie, and Zhang]{wang2022usb}
Yidong Wang, Hao Chen, Yue Fan, Wang Sun, Ran Tao, Wenxin Hou, Renjie Wang,
  Linyi Yang, Zhi Zhou, Lan-Zhe Guo, Heli Qi, Zhen Wu, Yu-Feng Li, Satoshi
  Nakamura, Wei Ye, Marios Savvides, Bhiksha Raj, Takahiro Shinozaki, Bernt
  Schiele, Jindong Wang, Xing Xie, and Yue Zhang.
\newblock {USB}: A unified semi-supervised learning benchmark for
  classification.
\newblock In \emph{Adv. Neural Inf. Process. Syst. (NeurIPS)},
  2022{\natexlab{a}}.

\bibitem[Wang et~al.(2022{\natexlab{b}})Wang, Wang, Shen, Fei, Li, Jin, Wu,
  Zhao, and Le]{wang2022u2pl}
Yuchao Wang, Haochen Wang, Yujun Shen, Jingjing Fei, Wei Li, Guoqiang Jin,
  Liwei Wu, Rui Zhao, and Xinyi Le.
\newblock Semi-supervised semantic segmentation using unreliable pseudo-labels.
\newblock In \emph{Proc. IEEE/CVF Conf. Comput. Vis. Pattern Recognit. (CVPR)},
  pages 4248--4257, 2022{\natexlab{b}}.

\bibitem[Wang et~al.(2023)Wang, Chen, Heng, Hou, Fan, Wu, Wang, Savvides,
  Shinozaki, Raj, Schiele, and Xie]{wang2023freematch}
Yidong Wang, Hao Chen, Qiang Heng, Wenxin Hou, Yue Fan, Zhen Wu, Jindong Wang,
  Marios Savvides, Takahiro Shinozaki, Bhiksha Raj, Bernt Schiele, and Xing
  Xie.
\newblock {FreeMatch}: Self-adaptive thresholding for semi-supervised learning.
\newblock In \emph{Int. Conf. Learn. Represent. (ICLR)}, 2023.

\bibitem[Wu et~al.(2023)Wu, Yang, Gan, Ding, Jiang, and Nie]{wu2023chmatch}
Jianlong Wu, Haozhe Yang, Tian Gan, Ning Ding, Feijun Jiang, and Liqiang Nie.
\newblock {CHMatch}: Contrastive hierarchical matching and robust adaptive
  threshold boosted semi-supervised learning.
\newblock In \emph{Proc. IEEE/CVF Conf. Comput. Vis. Pattern Recognit. (CVPR)},
  pages 15762--15772, 2023.

\bibitem[Xie et~al.(2020)Xie, Dai, Hovy, Luong, and Le]{xie2020self}
Qizhe Xie, Zihang Dai, Eduard Hovy, Thang Luong, and Quoc~V. Le.
\newblock Unsupervised data augmentation for consistency training.
\newblock In \emph{Adv. Neural Inf. Process. Syst. (NeurIPS)}, 2020.

\bibitem[Yang et~al.(2023{\natexlab{a}})Yang, Shi, Wei, Liu, Zhao, Ke, Pfister,
  and Ni]{medmnist2021}
Jiancheng Yang, Rui Shi, Donglai Wei, Zequan Liu, Lin Zhao, Bilian Ke,
  Hanspeter Pfister, and Bingbing Ni.
\newblock {MedMNIST} v2---a large-scale lightweight benchmark for {2D} and {3D}
  biomedical image classification.
\newblock \emph{Scientific Data}, 10\penalty0 (1):\penalty0 41,
  2023{\natexlab{a}}.

\bibitem[Yang et~al.(2023{\natexlab{b}})Yang, Qi, Feng, Zhang, and
  Shi]{yang2023unimatch}
Lihe Yang, Lei Qi, Litong Feng, Wayne Zhang, and Yinghuan Shi.
\newblock Revisiting weak-to-strong consistency in semi-supervised semantic
  segmentation.
\newblock In \emph{Proc. IEEE/CVF Conf. Comput. Vis. Pattern Recognit. (CVPR)},
  pages 7236--7246, 2023{\natexlab{b}}.

\bibitem[Yang et~al.(2025)Yang, Zhao, and Zhao]{yang2025unimatchv2}
Lihe Yang, Zhen Zhao, and Hengshuang Zhao.
\newblock {UniMatch V2}: Pushing the limit of semi-supervised semantic
  segmentation.
\newblock \emph{IEEE Transactions on Pattern Analysis and Machine
  Intelligence}, 47\penalty0 (4):\penalty0 3031--3048, 2025.

\bibitem[Yun et~al.(2019)Yun, Han, Oh, Chun, Choe, and Yoo]{yun2019cutmix}
Sangdoo Yun, Dongyoon Han, Seong~Joon Oh, Sanghyuk Chun, Junsuk Choe, and
  Youngjoon Yoo.
\newblock {CutMix}: Regularization strategy to train strong classifiers with
  localizable features.
\newblock In \emph{Proc. IEEE/CVF Int. Conf. Comput. Vis. (ICCV)}, pages
  6023--6032, 2019.

\bibitem[Zeng et~al.(2023)Zeng, Xie, Lu, and Xia]{zeng2023pefat}
Qingjie Zeng, Yutong Xie, Zilin Lu, and Yong Xia.
\newblock {PEFAT}: Boosting semi-supervised medical image classification via
  pseudo-loss estimation and feature adversarial training.
\newblock In \emph{Proc. IEEE/CVF Conf. Comput. Vis. Pattern Recognit. (CVPR)},
  pages 15671--15680, 2023.

\bibitem[Zhang et~al.(2021)Zhang, Wang, Hou, Wu, Wang, Okumura, and
  Shinozaki]{zhang2021flexmatch}
Bowen Zhang, Yidong Wang, Wenxin Hou, Hao Wu, Jindong Wang, Manabu Okumura, and
  Takahiro Shinozaki.
\newblock {FlexMatch}: Boosting semi-supervised learning with curriculum pseudo
  labeling.
\newblock In \emph{Adv. Neural Inf. Process. Syst. (NeurIPS)}, 2021.

\bibitem[Zhang et~al.(2023)Zhang, Li, Gao, and Giannakis]{zhang2023ibaml}
Yilang Zhang, Bingcong Li, Shijian Gao, and Georgios~B. Giannakis.
\newblock Scalable bayesian meta-learning through generalized implicit
  gradients.
\newblock In \emph{Proc. AAAI Conf. Artif. Intell.}, pages 11298--11306, 2023.

\bibitem[Zhao et~al.(2026)Zhao, Li, Bai, Shao, and Li]{zhao2026iasft}
Keyun Zhao, Guangchang Li, Yunpeng Bai, Jiang Shao, and Ying Li.
\newblock Rethinking low-confidence pseudo labels: Influence-aware
  semi-supervised fine-tuning for hyperspectral change detection.
\newblock In \emph{Proc. Int. Conf. Mach. Learn. (ICML)}, pages 161608--161622,
  2026.

\bibitem[Zheng et~al.(2022)Zheng, You, Huang, Wang, Qian, and
  Xu]{zheng2022simmatch}
Mingkai Zheng, Shan You, Lang Huang, Fei Wang, Chen Qian, and Chang Xu.
\newblock {SimMatch}: Semi-supervised learning with similarity matching.
\newblock In \emph{Proc. IEEE/CVF Conf. Comput. Vis. Pattern Recognit. (CVPR)},
  pages 14471--14481, 2022.

\bibitem[Zhou et~al.(2017)Zhou, Zhao, Puig, Fidler, Barriuso, and
  Torralba]{zhou2017ade20k}
Bolei Zhou, Hang Zhao, Xavier Puig, Sanja Fidler, Adela Barriuso, and Antonio
  Torralba.
\newblock Scene parsing through {ADE20K} dataset.
\newblock In \emph{Proc. IEEE Conf. Comput. Vis. Pattern Recognit. (CVPR)},
  pages 633--641, 2017.

\bibitem[Zhou et~al.(2023)Zhou, Lu, Liu, Xu, Cheng, and
  Niu]{zhou2023hypermatch}
Beitong Zhou, Jing Lu, Kerui Liu, Yunlu Xu, Zhanzhan Cheng, and Yi Niu.
\newblock {HyperMatch}: Noise-tolerant semi-supervised learning via relaxed
  contrastive constraint.
\newblock In \emph{Proc. IEEE/CVF Conf. Comput. Vis. Pattern Recognit. (CVPR)},
  pages 24017--24026, 2023.

\end{thebibliography}
}

% Supplementary material follows the paper and its consolidated bibliography.
% Defining MainOnly produces the paper alone for a separate preview.
\ifdefined\MainOnly\else
\clearpage
\renewcommand{\topfraction}{0.95}
\renewcommand{\dbltopfraction}{0.95}
\renewcommand{\textfraction}{0.05}
\renewcommand{\floatpagefraction}{0.75}
\renewcommand{\dblfloatpagefraction}{0.75}
\setcounter{topnumber}{3}
\setcounter{dbltopnumber}{3}
\setlength{\textfloatsep}{9pt plus 2pt minus 2pt}
\setlength{\dbltextfloatsep}{10pt plus 2pt minus 2pt}
\setlength{\floatsep}{8pt plus 2pt minus 2pt}
\setlength{\dblfloatsep}{8pt plus 2pt minus 2pt}

\makeatletter
\setlength{\@fptop}{0pt}
\setlength{\@dblfptop}{0pt}
\setlength{\@dblfpsep}{8pt plus 1pt minus 1pt}
\makeatother

\captionsetup{hypcap=false}
\captionsetup[table]{skip=5pt}
\captionsetup[figure]{skip=5pt}
% Supplementary material for the public preprint.
\setcounter{section}{0}
\setcounter{table}{0}
\setcounter{figure}{0}
\setcounter{equation}{0}
\renewcommand{\thesection}{S\arabic{section}}
\renewcommand{\thetable}{S\arabic{table}}
\renewcommand{\thefigure}{S\arabic{figure}}
\renewcommand{\theequation}{S\arabic{equation}}
\renewcommand{\theHsection}{supp.\arabic{section}}
\renewcommand{\theHtable}{supp.\arabic{table}}
\renewcommand{\theHfigure}{supp.\arabic{figure}}
\renewcommand{\theHequation}{supp.\arabic{equation}}
\twocolumn[{
\begin{center}
\Large\bfseries Rescuing Discarded Boundary Samples in Semi-Supervised Learning\par
\vspace{5pt}
\large\normalfont Supplementary Material\par
\end{center}
\vspace{6pt}
}]
\raggedbottom
\setlength{\abovedisplayskip}{7pt plus 2pt minus 2pt}
\setlength{\belowdisplayskip}{7pt plus 2pt minus 2pt}
\setlength{\belowdisplayshortskip}{5pt plus 1pt minus 1pt}

Sections~\ref{sec:supp-config}--\ref{sec:supp-classification-settings} give the
classification settings, with full seed-level results in
Section~\ref{sec:supp-main-seeds}. Section~\ref{sec:supp-pixelbrc} describes
PixelBRC. Sections~\ref{sec:supp-analysis}--\ref{sec:supp-additional-cases}
give the analysis protocols, additional quantitative evidence, and case studies.

The accompanying code package includes the full BRC implementation for the main
classification experiments, separate configurations for the main-paper component
ablations, and the PixelBRC implementation for semantic segmentation; installation
and usage instructions are in the README.

\Needspace{5\baselineskip}
\section{Datasets and training configuration}
\label{sec:supp-config}
We describe the classification data, backbones, augmentation, and shared
optimization settings.

\subsection{Data, backbones, and augmentation}
We use the six classification settings from the main paper: CIFAR-100 with
100/200 labels,
STL-10 with 40/100 labels, and TissueMNIST with 100/200 labels. The class
counts are 100, 10, and 8, respectively. The main classification comparisons
and component ablations use seeds 1, 2, and 42; Base and BRC share the labeled
split and unlabeled pool within each seed. Their classification errors are
reported as mean and standard deviation over these three runs. The single-seed
diagnostic comparisons and hyperparameter sweeps are described separately below.

\begin{table}[!htbp]
\caption{Classification architecture and dataset-specific configuration.
The number after the slash denotes the patch size.}
\label{tab:supp-data-config}
\centering\footnotesize
\setlength{\tabcolsep}{3pt}
\begin{tabular}{@{}lccc@{}}
\toprule
Setting & CIFAR-100 & STL-10 & TissueMNIST\\
\midrule
Backbone & ViT-S/2 & ViT-B/16 & ViT-Ti/2\\
Input resolution & $32^2$ & $96^2$ & $32^2$\\
Feature dimension $d$ & 384 & 768 & 192\\
Attention heads & 6 & 12 & 3\\
Transformer blocks & 12 & 12 & 12\\
Backbone LR & $5\!\times\!10^{-4}$ & $10^{-4}$ & $5\!\times\!10^{-5}$\\
Layer LR decay & 0.50 & 0.65 & 0.95\\
Crop ratio & 0.875 & 0.875 & 0.95\\
\bottomrule
\end{tabular}
\end{table}

Table~\ref{tab:supp-data-config} lists the dataset-specific backbone settings.
We follow the USB~\cite{wang2022usb} setup for pretrained initialization. The small and tiny
ViTs use the corresponding ImageNet-1K checkpoints, and ViT-B/16 uses the
MAE-pretrained checkpoint~\cite{he2022mae}. Stochastic-depth rates are 0.2 for ViT-S/2 and
ViT-B/16 and 0.1 for ViT-Ti/2.

We use the same augmentation pipeline for Base and BRC. Weak views use
resizing, random cropping with
reflection padding, and horizontal flipping. Strong views additionally use
RandAugment~\cite{cubuk2020randaugment} with three operations and magnitude five. TissueMNIST is resized
to $32\times32$ and converted to RGB; its transforms also include vertical
flipping and exclude color augmentation from RandAugment. Evaluation uses
resizing and dataset normalization without stochastic augmentation.

\subsection{Optimization settings}
We use the RCN architecture specified in the main paper.

All classification runs use 204,800 iterations and minibatches of 32 labeled
and 32 unlabeled images. Evaluation uses batch size 64 every 2,048 iterations. The backbone optimizer is AdamW with weight decay
$5\times10^{-4}$ and default $\beta=(0.9,0.999)$, using the learning rates and
layer decay in Table~\ref{tab:supp-data-config}.

The learning rate follows a cosine schedule. Its multiplier is
$\cos(7\pi t/16)$ after linear warm-up, where $t\in[0,1]$ is normalized
training progress after warm-up. FixMatch and FlexMatch use 5,120 LR warm-up iterations for STL-10
with 40 labels and none for their other settings. FreeMatch,
OTMatch, and RegMixMatch use 5,120. Mixed precision and gradient clipping are
disabled, and we do not use an EMA backbone for classification. RCN uses a separate
Adam optimizer with LR $5\times10^{-4}$ and score-loss weight 0.1.

\Needspace{5\baselineskip}
\section{Classification settings and seed-level results}
\label{sec:supp-classification-settings}
We follow the BRC formulation in the main paper, including
\emph{rescue-augmented backbone update} and \emph{validation-guided RCN update}.
Here we list the base-learner hyperparameters and the schedules used in the
classification experiments.

\subsection{Base-learner hyperparameters}
Hard pseudo-label targets use temperature setting 0.5. FixMatch and
FlexMatch use unsupervised CE weight one. FixMatch uses a confidence
threshold of 0.95.
FlexMatch uses a nominal cutoff of 0.95 with threshold warm-up enabled.
FreeMatch retains its fairness/entropy weight of 0.001 and
probability-statistics EMA of 0.999. OTMatch uses entropy weight 0.001,
transport-loss weight 0.15, and class-cost EMA 0.999.

RegMixMatch uses entropy weight 0.001, ResizeMix weight 1.0, and class-aware
mixing weight 1.0. Its mixing settings are $\tau_m=0.99$,
$\alpha_h=1$, $\alpha_l=16$, class-aware mix probability 0.5, and confidence
power 3. Base and BRC use the same base-learner settings within each
experimental pair.

\subsection{BRC schedules and rescue ratios}
Table~\ref{tab:supp-schedule} lists the start iterations and rescue ratios.
These settings are held fixed across datasets and label budgets for each
base learner. RCN updates use an interval of $K_{\rm rcn}=10$ iterations.

\begin{table}[!htbp]
\caption{Classification BRC experiment settings. $T_{\rm rcn}$ and
$T_{\rm res}$ denote the start iterations for RCN updates and rescue;
$\rho$ is the rescue ratio.}
\label{tab:supp-schedule}
\centering\footnotesize
\setlength{\tabcolsep}{3pt}
\begin{tabular}{@{}lrrcl@{}}
\toprule
Base & $T_{\rm rcn}$ & $T_{\rm res}$ & $\rho$ & RCN cutoff\\
\midrule
FixMatch & 50,000 & 55,000 & 20\% & 0.95\\
FlexMatch & 0 & 5,000 & 30\% & 0.98\\
FreeMatch & 50,000 & 55,000 & 50\% & Adaptive\\
OTMatch & 50,000 & 55,000 & 50\% & 0.95\\
RegMixMatch & 0 & 5,000 & 30\% & 0.98\\
\bottomrule
\end{tabular}
\end{table}

\subsection{Full seed-level classification results}
\label{sec:supp-main-seeds}
Table~\ref{tab:supp-main-seeds} gives all retained seed-level records underlying
main-paper Table~1. Entries are Top-1 error rates, computed as
$100(1-a)$ from the archived accuracy $a$ and displayed to three decimal places.

\begin{table*}[!t]
\caption{Full seed-level classification results underlying main-paper Table~1.
Entries are Top-1 error rates (\%); lower is better. Parentheses give the label
budget. Base and BRC share the data split within each paired setting and seed.}
\label{tab:supp-main-seeds}
\centering\footnotesize
\setlength{\tabcolsep}{2pt}
\renewcommand{\arraystretch}{1.06}
\begin{tabular}{@{}l*{12}{r}@{}}
\toprule
\textbf{CIFAR-100} & \multicolumn{3}{c}{Base (100)} & \multicolumn{3}{c}{BRC (100)} & \multicolumn{3}{c}{Base (200)} & \multicolumn{3}{c}{BRC (200)} \\
\cmidrule(lr){2-4}\cmidrule(lr){5-7}\cmidrule(lr){8-10}\cmidrule(lr){11-13}
Method & Seed 1 & Seed 2 & Seed 42 & Seed 1 & Seed 2 & Seed 42 & Seed 1 & Seed 2 & Seed 42 & Seed 1 & Seed 2 & Seed 42 \\
\midrule
FixMatch & 53.710 & 54.210 & 51.640 & 52.670 & 53.120 & 52.420 & 31.540 & 35.410 & 27.700 & 32.420 & 31.350 & 28.140 \\
FlexMatch & 39.510 & 43.520 & 45.000 & 34.050 & 38.500 & 42.950 & 28.280 & 27.570 & 25.400 & 25.180 & 22.810 & 21.560 \\
FreeMatch & 33.500 & 39.490 & 41.250 & 33.220 & 34.410 & 37.200 & 22.730 & 24.180 & 17.410 & 22.370 & 20.090 & 18.800 \\
OTMatch & 38.580 & 43.430 & 42.430 & 33.640 & 35.880 & 40.020 & 21.960 & 24.040 & 18.690 & 22.470 & 20.420 & 18.550 \\
RegMixMatch & 34.170 & 44.130 & 40.340 & 32.630 & 42.280 & 39.170 & 25.680 & 24.460 & 16.410 & 24.880 & 22.370 & 15.720 \\
\midrule
\textbf{STL-10} & \multicolumn{3}{c}{Base (40)} & \multicolumn{3}{c}{BRC (40)} & \multicolumn{3}{c}{Base (100)} & \multicolumn{3}{c}{BRC (100)} \\
\cmidrule(lr){2-4}\cmidrule(lr){5-7}\cmidrule(lr){8-10}\cmidrule(lr){11-13}
Method & Seed 1 & Seed 2 & Seed 42 & Seed 1 & Seed 2 & Seed 42 & Seed 1 & Seed 2 & Seed 42 & Seed 1 & Seed 2 & Seed 42 \\
\midrule
FixMatch & 16.888 & 23.425 & 16.150 & 10.250 & 22.412 & 10.875 & 4.850 & 6.300 & 6.875 & 4.312 & 4.862 & 5.650 \\
FlexMatch & 7.800 & 16.612 & 18.012 & 10.088 & 16.325 & 14.975 & 5.375 & 7.875 & 6.412 & 5.062 & 6.725 & 6.812 \\
FreeMatch & 9.438 & 13.612 & 15.288 & 8.238 & 10.425 & 12.988 & 5.500 & 7.000 & 6.050 & 5.138 & 6.750 & 6.575 \\
OTMatch & 10.725 & 13.150 & 15.900 & 8.025 & 12.975 & 16.988 & 5.462 & 7.200 & 6.425 & 5.125 & 6.375 & 7.075 \\
RegMixMatch & 6.550 & 14.900 & 13.150 & 7.738 & 8.888 & 7.762 & 5.312 & 8.050 & 6.588 & 5.588 & 7.250 & 6.288 \\
\midrule
\textbf{TissueMNIST} & \multicolumn{3}{c}{Base (100)} & \multicolumn{3}{c}{BRC (100)} & \multicolumn{3}{c}{Base (200)} & \multicolumn{3}{c}{BRC (200)} \\
\cmidrule(lr){2-4}\cmidrule(lr){5-7}\cmidrule(lr){8-10}\cmidrule(lr){11-13}
Method & Seed 1 & Seed 2 & Seed 42 & Seed 1 & Seed 2 & Seed 42 & Seed 1 & Seed 2 & Seed 42 & Seed 1 & Seed 2 & Seed 42 \\
\midrule
FixMatch & 54.824 & 53.739 & 56.865 & 52.648 & 53.581 & 56.337 & 51.990 & 53.190 & 54.412 & 51.442 & 51.802 & 55.398 \\
FlexMatch & 57.473 & 56.244 & 59.816 & 54.526 & 55.755 & 59.416 & 51.933 & 54.133 & 57.392 & 52.745 & 52.802 & 56.360 \\
FreeMatch & 56.307 & 55.345 & 60.493 & 55.027 & 54.480 & 57.475 & 51.965 & 52.155 & 57.358 & 51.703 & 53.354 & 56.383 \\
OTMatch & 55.370 & 55.533 & 60.144 & 55.573 & 54.306 & 57.071 & 51.703 & 53.420 & 56.937 & 50.641 & 53.058 & 55.097 \\
RegMixMatch & 57.168 & 55.241 & 57.625 & 54.915 & 54.600 & 55.785 & 51.385 & 52.824 & 53.412 & 49.459 & 53.236 & 53.147 \\
\bottomrule
\end{tabular}
\end{table*}

\Needspace{5\baselineskip}
\section{PixelBRC implementation details}
\label{sec:supp-pixelbrc}
PixelBRC makes a rescue decision for each pixel. We describe its dense
module and spatial selection rule, then explain labeled supervision and
target alignment. Training configurations and base-method differences follow.

\subsection{Dense base process and module inputs}
PixelBRC uses teacher predictions and decoder features to correct rejected
pixels. Let $P_v=f_{\bar\theta}(u^w)_v\in\Delta^{C-1}$ be the EMA teacher prediction
at pixel $v$, $c_v=\max_k P_{vk}$, and $\widetilde y_v=\arg\max_k P_{vk}$.
The base method provides $m_v=\mathbb{I}[c_v\geq\tau_v]$ on valid pixels.
Here $\bar\theta$ denotes the EMA teacher, $C$ the class count, and
$\tau_v$ the base learner's fixed or adaptive threshold. The student uses
labeled CE and masked pseudo-label CE on strong views. PixelBRC adds targets
only for selected rejected pixels.

We use DINOv2-Base~\cite{oquab2024dinov2} with a DPT decoder~\cite{ranftl2021dpt}. The module reads the decoder's final
128-channel feature map $H$ at the prediction resolution. PixelBRC takes
$[H_v,P_v,c_v,e_v]$, where
$e_v=-\sum_kP_{vk}\log P_{vk}/\log C$ is normalized entropy. A $1\times1$
convolution projects the $128+C+2$ input channels to 128, followed by
BatchNorm and ReLU; a $3\times3$ convolution with padding one, BatchNorm, and
ReLU provides spatial context. Two $1\times1$ heads produce correction
logits $a_v\in\mathbb{R}^C$ and score logits $b_v\in\mathbb{R}$. The selection
score is $s_v=\sigma(b_v)$. We add the correction logits to the teacher's log
probabilities and apply softmax to obtain
\begin{equation}
Q_v=\operatorname{softmax}\bigl(\log(\max(P_v,10^{-6}))+a_v\bigr).
\label{eq:pixel-residual}
\end{equation}
This retains the teacher distribution as a reference while allowing a learned
class-dependent change.

\subsection{Spatial prior and rejected-pixel selection}
Nearby confident predictions guide rescue at semantic boundaries. We first
build a local class prior, then use selection scores and neighborhood support to
select rejected pixels.

A predicted boundary pixel has a different pseudo-label from at least one
valid four-connected neighbor. We dilate this mask by one pixel to obtain
$B_v$. Let $\mathcal{N}_3(v)$ be the $7\times7$ neighborhood of $v$.
Using high-confidence teacher pseudo-labels, we compute class counts,
support, and a local prior:
\begin{equation}
\begin{aligned}
n_{vk}&=\sum_{w\in\mathcal{N}_3(v)}m_w\mathbb{I}[\widetilde y_w=k],\\
a_v^{\rm nbr}&=\frac{\sum_k n_{vk}}{49},\qquad
\pi_{vk}=\frac{n_{vk}}{\max(\sum_k n_{vk},10^{-6})}.
\end{aligned}
\label{eq:pixel-prior}
\end{equation}
Here $n_{vk}$ counts accepted neighbors of class $k$, $a_v^{\rm nbr}$ is
their support fraction, and $\pi_v$ is the local class distribution. We use
average pooling with zero padding to compute these quantities.

We next restrict rescue to rejected pixels with moderate confidence near a
predicted boundary. The initial candidate set in each image is
\begin{equation}
\mathcal C=\{v:0.30\leq c_v<\tau_v,\ B_v=1,\ a_v^{\rm nbr}\geq0.05\}.
\end{equation}
If this set is empty in an image, we fall back to all pixels with
$0.30\leq c_v<\tau_v$ in that image. We rank candidates by
$s_v a_v^{\rm nbr}$ and take the top
$\max(1,\lfloor0.10|\mathcal C|\rfloor)$ when $\mathcal C$ is nonempty.
We retain a ranked candidate only if its selection score, corrected
confidence, and local support for the corrected class satisfy
\begin{equation}
s_v\geq0.50,\quad \max_kQ_{vk}\geq0.60,\quad
\pi_{v,\arg\max_kQ_{vk}}\geq0.10.
\label{eq:pixel-gates}
\end{equation}
The rescue indicator $r_v$ is one only for pixels passing all three checks.
Selection is per image, and ignore pixels are excluded from training.

For each selected pixel, we form a soft target by combining the teacher
prediction, learned correction, and local class prior:
\begin{equation}
\widehat Y_v=(1-\beta-\gamma_v)P_v+\beta Q_v+\gamma_v\pi_v,
\label{eq:pixel-target}
\end{equation}
where $\beta=0.50$ and $\gamma_v=0.25$ when neighborhood support is positive,
and zero otherwise. The result is normalized along the class dimension.
Confidence gates in Eq.~\eqref{eq:pixel-gates} are applied to $Q_v$ before
this blend.

\subsection{Labeled supervision}
Both heads receive direct supervision from labeled pixels, with training
focused on uncertain regions and semantic boundaries. The module receives
detached student features and probabilities and applies the residual
correction in Eq.~\eqref{eq:pixel-residual}. Valid pixels within a two-pixel
dilation of a ground-truth boundary or with prediction confidence below 0.95
form the labeled training set.

The correction head uses cross-entropy against the ground-truth class.
The score head uses confidence-based binary supervision reflecting the
original prediction's support for that class, with binary cross-entropy
weight 0.1. The labeled module loss combines the correction and score terms.
The score guides pixel selection together with the spatial and confidence
criteria described above. The supervision targets are fixed when
optimizing the module, and detached inputs restrict these losses to PixelBRC.
If the labeled training set is empty, its module loss is zero. Ground-truth
boundaries are used only on labeled images; unlabeled rescue uses predicted
boundaries.

\subsection{View alignment and student objective}
To align each corrected target with its student prediction, we transform
the soft targets and rescue masks together.

Targets are constructed from the teacher's weak view without gradients.
The weak and strong views share the geometric resize, crop, and flip; strong
views add color jitter, grayscale, and blur. For CutMix~\cite{yun2019cutmix}, let $M_v$ select a
pixel from another image in the minibatch, indexed by $\kappa$. We transform both target and
rescue mask using the exact strong-view mixing box:
\begin{equation}
\begin{aligned}
\widehat Y^{\rm mix}_v&=(1-M_v)\widehat Y_v+M_v\widehat Y_{\kappa,v},\\
r^{\rm mix}_v&=(1-M_v)r_v+M_vr_{\kappa,v}.
\end{aligned}
\end{equation}
The ignore mask is mixed identically and intersected with $r^{\rm mix}$.
Each strong-view rescue loss is mean soft CE over its selected valid pixels:
\begin{equation}
\mathcal L_{{\rm pix},u}^{(q)}=
-\frac{\sum_v r_v^{(q)}\sum_k\widehat Y_{vk}^{(q)}
\log f_\theta(u^{s_q})_{vk}}{\max(1,\sum_vr_v^{(q)})}.
\end{equation}
Here $q$ indexes the strong view, and the sums include the minibatch.
For $V$ strong views, the rescue loss is
$\mathcal L_{{\rm pix},u}=V^{-1}\sum_{q=1}^{V}\mathcal L_{{\rm pix},u}^{(q)}$.
The original accepted-pixel loss $\mathcal L_u$ is normalized by all valid
pixels in each view and averaged over the same $V$ views. The full objective is
\begin{equation}
\mathcal L=\frac{\mathcal L_{\rm sup}+\mathcal L_u}{2}
+0.05\mathcal L_{{\rm pix},\ell}
+\lambda_{\rm pix}(t)\mathcal L_{{\rm pix},u}.
\label{eq:pixel-total}
\end{equation}
Here $\lambda_{\rm pix}(t)=0$ before rescue starts and 0.05 thereafter.
The labeled module loss is active from the beginning. The unlabeled rescue
loss trains the student, and its detached targets do not train PixelBRC.
Both PixelBRC and the teacher are discarded for student inference. The
segmentation results evaluate the complete extension, including its
spatial selection, correction, and target construction.

\subsection{Shared segmentation configuration}
Base and PixelBRC share data splits, networks, and optimizer settings.
We give these settings below; Table~\ref{tab:supp-seg-config} lists the
dataset-specific training schedules.

The labeled/unlabeled image counts are 92/275 for CamVid, 183/10,399
for VOC with SBD augmentation~\cite{hariharan2011semantic}, 1,263/18,947 for ADE20K, and 143/429 for
Stanford Background. The VOC 1/8 budget is relative to the 1,464 original
training images; the unlabeled pool includes SBD augmentation. Labeled images
are repeated to match the unlabeled epoch length. Geometric augmentation uses
random rescaling in $[0.5,2.0]$, cropping, and horizontal flipping.

All settings use the same pretrained DINOv2-Base/DPT architecture, with the
encoder unfrozen and $518\times518$ training crops. AdamW uses
$\beta=(0.9,0.999)$ and weight decay 0.01. The encoder LR is $5\times10^{-6}$;
the decoder and PixelBRC LR is $2\times10^{-4}$ (multiplier 40), with polynomial
decay $(1-t/T)^{0.9}$. The teacher EMA coefficient at step $t$ is
$\min(1-1/(t+1),0.996)$. Ignore label 255 is excluded from all training losses.
We report best validation student mIoU for each matched run.

\begin{table}[!htbp]
\caption{PixelBRC training configuration. Batch denotes the size of each
labeled/unlabeled minibatch; Start is the first rescue iteration. Splits
are CamVid/Stanford BG 1/4, VOC 1/8, and ADE20K 1/16.}
\label{tab:supp-seg-config}
\centering\footnotesize\setlength{\tabcolsep}{3pt}
\begin{tabular}{@{}llrrr@{}}
\toprule
Dataset & Base & Epochs & Batch & Start\\
\midrule
CamVid & Dense FreeMatch & 120 & 4 & 500\\
VOC & FixMatch & 60 & 4 & 2,000\\
ADE20K & UniMatch V2 & 60 & 8 & 2,000\\
Stanford BG & Teacher--student & 80 & 2 & 500\\
CamVid & FixMatch & 120 & 4 & 500\\
\bottomrule
\end{tabular}
\end{table}

\subsection{Integration with base learners}
\label{sec:supp-pixel-base-integration}
The base learners use different acceptance thresholds and strong views
(Table~\ref{tab:supp-pixel-base-integration}), while sharing the PixelBRC
module, supervision, rescue rule, and loss weights.

\begin{table}[!htbp]
\caption{Base-specific PixelBRC integration. $V$ is the number of strong
views; accepted-pixel and rescue losses are averaged over these views.}
\label{tab:supp-pixel-base-integration}
\centering\footnotesize\setlength{\tabcolsep}{3pt}
\renewcommand{\arraystretch}{1.12}
\begin{tabular}{@{}lcc>{\raggedright\arraybackslash}p{0.38\columnwidth}@{}}
\toprule
Base learner & Threshold & $V$ & Strong-view training\\
\midrule
FixMatch & Fixed 0.95 & 1 & CutMix\\
Dense FreeMatch & Adaptive & 1 & CutMix\\
UniMatch V2 & Fixed 0.95 & 2 & CutMix and complementary feature dropout\\
Teacher--student & Fixed 0.95 & 1 & CutMix disabled\\
\bottomrule
\end{tabular}
\end{table}

Dense FreeMatch computes a class-dependent cutoff by scaling its EMA mean
confidence by the class's mean probability relative to the largest class
mean probability. We use EMA momentum 0.99 and clip cutoffs to the range
0.50--0.95. The threshold map receives the same CutMix transform as the
pseudo-labels and is used consistently for rejection and neighborhood-prior
construction. Labeled PixelBRC supervision uses the nominal confidence
threshold 0.95 for all four bases.

\Needspace{5\baselineskip}
\section{Boundary analysis protocols}
\label{sec:supp-analysis}
We specify the feature-distance calculation, checkpoint replay, and fixed-set
evaluation used in the main-paper boundary analyses.

\subsection{Distribution analysis of rejected samples}
The centroid-distance ratio compares a feature's distances to its true class
and nearest competing class. We use all 50,000 CIFAR-100 training images at
the midpoint FixMatch checkpoint, keeping image order consistent across
features, labels, and acceptance masks. Let
$h_i$ be a backbone feature and $\mathcal I_k=\{i:y_i=k\}$. Feature
normalization precedes class averaging, and each class mean is normalized
again:
\begin{equation}
\begin{aligned}
z_i&=\frac{h_i}{\max(\|h_i\|_2,10^{-12})},\\
\bar z_k&=\frac{1}{|\mathcal I_k|}\sum_{i\in\mathcal I_k}z_i,\qquad
\mu_k=\frac{\bar z_k}{\max(\|\bar z_k\|_2,10^{-12})}.
\end{aligned}
\end{equation}
We form the cosine-distance matrix $D_{ik}=1-z_i^\top\mu_k$, exclude the
true-class column when finding the nearest competitor, and compute
\begin{equation}
R_i=\frac{D_{i,y_i}}{\max(\min_{k\ne y_i}D_{ik},10^{-8})}.
\end{equation}
Accepted and rejected groups use the saved model's confidence rule.
Ground-truth labels enter only the class-centroid diagnostic. The resulting
distributions and their interpretation are reported in the main paper.

\subsection{Rescue dynamics and boundary refinement}
Replay measures which examples the selector favors at a fixed checkpoint.
Base features define the boundary bands; the BRC checkpoint supplies its
rejection mask and utility scores from the RCN score head. We hold both models fixed and repeat
selection under different minibatch partitions. The full image pool is randomly permuted into
batches of 32; within each batch we rank only rejected examples and select
$\max(1,\lfloor0.30n_{\rm rej}\rfloor)$ when $n_{\rm rej}>0$.

Let $\mathcal S$ be the selected image IDs, $\mathcal B$ the Base-defined
boundary band, and $\mathcal U$ the full image pool. The reported rescue and
population fractions and their enrichment are
\begin{equation}
\begin{aligned}
f_{\rm res}&=\frac{|\mathcal S\cap\mathcal B|}{|\mathcal S|},\qquad
f_{\rm pool}=\frac{|\mathcal U\cap\mathcal B|}{|\mathcal U|},\\
E&=f_{\rm res}/f_{\rm pool}.
\end{aligned}
\end{equation}
We repeat the partitioning with 100 independent shuffles and summarize the
checkpoint-level selection statistics by their mean and standard deviation.
For the paired accuracy curves, both models are evaluated on the same
Base-defined sample IDs at each checkpoint.

\subsection{Comparison with lower-threshold SSL}
The threshold control uses FlexMatch with its nominal confidence threshold
lowered from 0.95 to 0.90, without BRC's learned rescue selection or label
correction.
Fixed sample sets allow us to compare models on the same examples even when
their predictions differ. We define the reference boundary sets once using
Base at iteration
73,728 and save their image IDs. Base, the threshold control, and BRC are
then evaluated at iterations
73,728, 145,408, and 204,800, respectively. Every model uses the saved IDs to define
boundary membership. The non-boundary set is the complement of
the main boundary band within the full training-image pool.

For an evaluation set $\mathcal A$ and model probabilities $P_i$, the metrics
are computed as
\begin{equation}
\begin{aligned}
\operatorname{Top1}(\mathcal A)
&=\frac{100}{|\mathcal A|}\sum_{i\in\mathcal A}
\mathbb I[\arg\max_k P_{ik}=y_i],\\
\operatorname{NLL}(\mathcal A)
&=-\frac{1}{|\mathcal A|}\sum_{i\in\mathcal A}
\log\max(P_{i,y_i},10^{-12}).
\end{aligned}
\end{equation}
Probabilities are obtained by softmax of each model's logits, using the same
sample ordering and reference subsets.

\Needspace{5\baselineskip}
\section{Additional quantitative evidence}
\label{sec:supp-sensitivity}
We examine the complementary roles of selection based on utility scores and
meta-learned correction through local gradient diagnostics and full-training
comparisons, followed by per-seed ablations and hyperparameter results.

\subsection{Local training value of selection and correction}
\label{sec:supp-rescue-value}
We use controlled comparisons to test whether selection based on utility scores
and meta-learned correction can provide useful supervision independently.
Using FlexMatch on CIFAR-100 with 200 labeled images, we compare rescue
selectors while retaining correction, and replace rescue targets while
retaining selection based on utility scores. At a fixed checkpoint, all comparisons share
the backbone, rejected pool, input views, validation examples, and total
rescue-loss weight. Budgeted selectors use the same number of samples;
target replacements reuse the same sample IDs selected by utility score.

For fixed rescue targets, let
$g_{\mathrm{res}}=\nabla_\theta\mathcal L_{\mathrm{res}}$.
We measure local training value by the first-order prediction of validation
loss reduction under a virtual step:
\begin{equation}
\begin{aligned}
\theta'&=\theta-\alpha g_{\mathrm{res}},\\
R_{\mathrm{first}}
&=\alpha\left\langle\nabla_\theta\mathcal L_{\mathrm{val}}(\theta),
g_{\mathrm{res}}\right\rangle.
\end{aligned}
\label{eq:supp-rescue-impact}
\end{equation}
Here $\alpha\approx4.72\times10^{-8}$, and positive
$R_{\mathrm{first}}$ predicts a validation-loss decrease. This is a local
gradient diagnostic rather than a measured final accuracy gain.

\begin{table}[!htbp]
\caption{Training value of rescued samples at a fixed checkpoint. Larger
$R_{\mathrm{first}}$ indicates a greater predicted decrease in held-out loss.}
\label{tab:supp-rescue-value}
\centering\footnotesize
\setlength{\tabcolsep}{3pt}
\begin{tabular}{@{}llr@{}}
\toprule
Selection & Rescue target & $R_{\mathrm{first}}$ ($10^{-8}$)\\
\midrule
Utility score & Learned correction & \textbf{3.21}\\
Original confidence & Learned correction & $-0.35$\\
Corrected confidence & Learned correction & 0.22\\
Random & Learned correction & $-0.42$\\
Utility score & Original soft prediction & 0.87\\
Utility score & Uniform soft label & $-2.11$\\
All rejected & Learned correction & $-1.50$\\
\bottomrule
\end{tabular}
\end{table}

Table~\ref{tab:supp-rescue-value} shows that correction without selection based on utility scores
yields weak or negative local gains, while such selection
without correction provides a smaller benefit than the combined strategy.
The strongest training value comes from coupling selection and correction,
supporting their complementary roles. We next test this conclusion through
full training.

\subsection{Full-training effects of selection and correction}
\label{sec:supp-rescue-training}
We train the corresponding variants using FlexMatch on CIFAR-100 with
200 labeled images, following the shared settings in
Sections~\ref{sec:supp-config}--\ref{sec:supp-classification-settings}.
The comparisons retain correction while replacing selection based on utility
scores, or retain this selection while replacing corrected targets. We match the
sample budget across budgeted selectors and the total rescue-loss weight
across rescue variants. All variants use a single shared random seed and
are trained independently from the same initialization and data split.

We report test accuracy and accuracy on a fixed narrow-boundary (Strict) set.
This set is defined once by the 200-label Base model using
$R_i\in[0.90,1.10]$ and the protocol in
Section~\ref{sec:supp-analysis}, and shared across configurations.

\begin{table}[!htbp]
\caption{Single-seed full-training comparison on CIFAR-100 with 200 labels
using FlexMatch. Each row reports test accuracy (\%) and accuracy (\%) on the
fixed narrow-boundary (Strict) set, $R_i\in[0.90,1.10]$.}
\label{tab:supp-rescue-training}
\centering\footnotesize
\setlength{\tabcolsep}{3pt}
\renewcommand{\arraystretch}{1.10}
\begin{tabular}{@{}lcc@{}}
\toprule
Configuration & Test acc. (\%) & \shortstack{Narrow-boundary\\acc. (\%)}\\
\midrule
Base & 72.41 & 30.57\\
All rejected + Correction & 71.96 & 32.65\\
Random + Correction & 72.24 & 33.39\\
Original confidence + Correction & 72.30 & 34.52\\
Corrected confidence + Correction & 70.98 & 34.03\\
Utility score + Original soft prediction & 73.36 & 32.04\\
Utility score + Uniform soft label & 68.13 & 25.88\\
Utility score + Correction & \textbf{77.50} & \textbf{44.53}\\
\bottomrule
\end{tabular}
\end{table}

Table~\ref{tab:supp-rescue-training} confirms the local evidence in
Section~\ref{sec:supp-rescue-value}. Meta-learned correction without
selection based on utility scores fails to improve test accuracy over Base and can reduce
it. Selection based on utility scores with the original targets gives only a modest gain,
whereas combining selection and correction achieves the best test and
narrow-boundary accuracies. Together, the two experiments show that neither
component alone recovers the full benefit of rescue in this adaptive-confidence
SSL setting: effective reuse requires selecting beneficial candidates and
providing suitable corrected supervision.

\subsection{Ablation results by seed}
Table~\ref{tab:supp-ablation-seeds} gives the CIFAR-100, 100-label ablations
for each seed. Shared variants follow the main-paper definitions. The additional ``w/o score loss''
variant sets the score-loss weight to zero while retaining the other BRC
components. The mean column is the average of the three seed accuracies. For the shared
variants, the main paper reports error as $100-\text{mean accuracy}$ and
uses the sample standard deviation across seeds.

\begin{table}[!htbp]
\caption{Per-seed Top-1 accuracy for component ablations. Seed values are
reported before aggregation; the mean column is their arithmetic mean.}
\label{tab:supp-ablation-seeds}
\centering\footnotesize
\setlength{\tabcolsep}{2pt}
\renewcommand{\arraystretch}{1.05}
\begin{tabular}{@{}llrrrr@{}}
\toprule
Base & Variant & Seed 1 & Seed 2 & Seed 42 & Mean (\%)\\
\midrule
FlexMatch & Base & 60.49 & 56.48 & 55.00 & 57.32\\
FlexMatch & Full BRC & 65.95 & 61.50 & 57.05 & 61.50\\
FlexMatch & Random selection & 58.67 & 58.49 & 54.11 & 57.09\\
FlexMatch & w/o correction & 57.81 & 56.32 & 50.25 & 54.79\\
FlexMatch & w/o score loss & 60.79 & 62.30 & 54.37 & 59.15\\
FreeMatch & Base & 66.50 & 60.51 & 58.75 & 61.92\\
FreeMatch & Full BRC & 66.78 & 65.59 & 62.80 & 65.06\\
FreeMatch & Random selection & 65.17 & 62.32 & 62.73 & 63.41\\
FreeMatch & w/o correction & 63.90 & 61.70 & 59.60 & 61.73\\
FreeMatch & w/o score loss & 65.31 & 61.35 & 62.37 & 63.01\\
RegMixMatch & Base & 65.83 & 55.87 & 59.66 & 60.45\\
RegMixMatch & Full BRC & 67.37 & 57.72 & 60.83 & 61.97\\
RegMixMatch & Random selection & 63.25 & 58.75 & 57.26 & 59.75\\
RegMixMatch & w/o correction & 56.44 & 52.97 & 57.07 & 55.49\\
RegMixMatch & w/o score loss & 63.17 & 57.30 & 58.13 & 59.53\\
\bottomrule
\end{tabular}
\end{table}

\subsection{Numerical results for BRC hyperparameters}
Table~\ref{tab:supp-sweep} provides the numerical results underlying the
hyperparameter-sensitivity figure in the main paper. We vary one parameter
at a time in FlexMatch on CIFAR-100 with 200 labels, using a single shared
random seed and keeping the other parameters fixed. Accuracy is the best validation Top-1 accuracy
recorded during training; the last column gives its iteration.

\begin{table}[!htbp]
\caption{Numerical results for the main-paper hyperparameter sensitivity
analysis. Entries are best validation Top-1 accuracies.}
\label{tab:supp-sweep}
\centering\footnotesize
\setlength{\tabcolsep}{3pt}
\renewcommand{\arraystretch}{1.00}
\begin{tabular}{@{}lrrr@{}}
\toprule
Parameter & Value & \shortstack{Best validation\\Top-1 (\%)} & \shortstack{Best\\iteration}\\
\midrule
$\tau_{\mathrm{rcn}}$ & 0.85 & 75.20 & 182,271\\
$\tau_{\mathrm{rcn}}$ & 0.90 & 78.34 & 122,879\\
$\tau_{\mathrm{rcn}}$ & 0.95 & 77.53 & 155,647\\
$\tau_{\mathrm{rcn}}$ & 0.98 & 77.50 & 188,415\\
$\tau_{\mathrm{rcn}}$ & 0.995 & 75.49 & 108,543\\
\midrule
$\rho$ (\%) & 10 & 75.89 & 190,463\\
$\rho$ (\%) & 20 & 75.68 & 182,271\\
$\rho$ (\%) & 30 & 77.50 & 188,415\\
$\rho$ (\%) & 40 & 78.28 & 190,463\\
$\rho$ (\%) & 50 & 76.44 & 92,159\\
\midrule
$T_{\mathrm{res}}$ & 0 & 77.00 & 172,031\\
$T_{\mathrm{res}}$ & 2,500 & 77.48 & 204,799\\
$T_{\mathrm{res}}$ & 5,000 & 77.50 & 188,415\\
$T_{\mathrm{res}}$ & 7,500 & 76.32 & 98,303\\
$T_{\mathrm{res}}$ & 10,000 & 74.56 & 204,799\\
\bottomrule
\end{tabular}
\end{table}

\Needspace{5\baselineskip}
\section{Additional qualitative case studies}
\label{sec:supp-additional-cases}
These cases supplement the main-paper examples.

\subsection{Image-classification case studies}
\label{sec:supp-class-cases}

The four CIFAR-100 cases in Figure~\ref{fig:supp-brc-cases} use the
main-paper FlexMatch+BRC checkpoint. We replay its acceptance mask and
select the top 30\% of rejected examples by utility score.

\begin{figure*}[!t]
\centering
\includegraphics[width=0.76\textwidth]{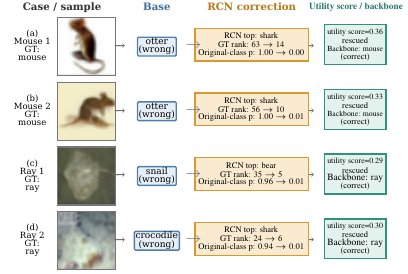}
\caption{Four additional CIFAR-100 cases: Base predictions, RCN targets,
rescue decisions, and final backbone predictions.}
\label{fig:supp-brc-cases}

\end{figure*}

\subsection{Semantic-segmentation case studies}
\label{sec:supp-seg-cases}
Figure~\ref{fig:supp-pixelbrc-cases} compares six additional VOC cases.
The images show local details; metrics use all valid pixels in a four-pixel
band around the full image's ground-truth boundary. Boundaries are identified
by label changes between four-connected neighbors. ``Improved'' counts
Base errors corrected by PixelBRC; ``degraded'' counts the reverse.
The gain is the boundary Top-1 difference in percentage points.

\begin{figure*}[!t]
\centering
\includegraphics[width=\textwidth]{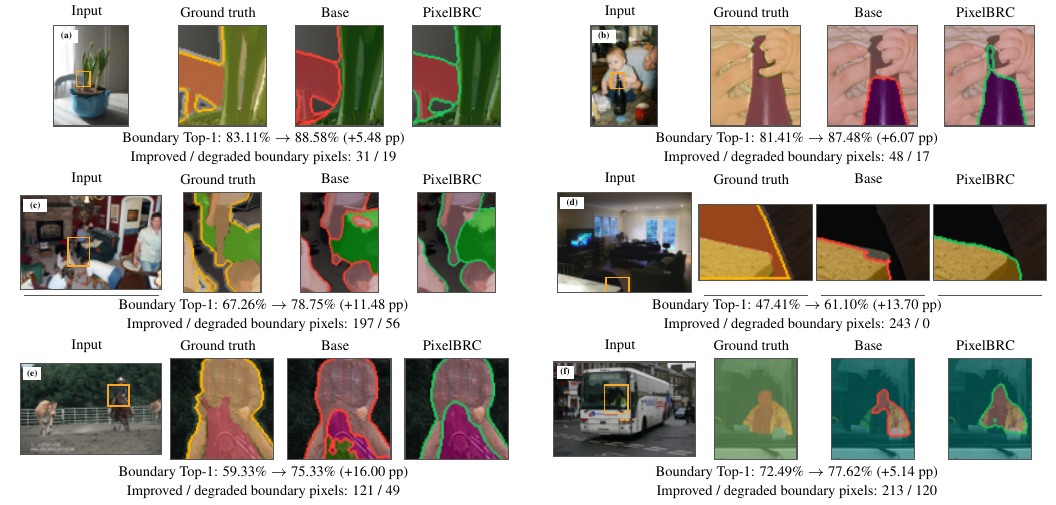}
\caption{Six additional VOC validation cases showing the input and zoomed
ground-truth, Base, and PixelBRC regions, with boundary metrics below. Yellow,
red, and green contours mark the respective boundaries.}
\label{fig:supp-pixelbrc-cases}

\end{figure*}

\clearpage

\fi

\end{document}